\documentclass[11pt]{article}

\usepackage[a4paper,margin=1in]{geometry}
\usepackage[T1]{fontenc}
\usepackage[utf8]{inputenc}
\usepackage{lmodern}
\usepackage{microtype}
\usepackage{amsmath,amssymb,amsthm,mathtools,bm}
\usepackage{booktabs}
\usepackage{enumitem}
\usepackage{natbib}
\usepackage{hyperref}
\usepackage{setspace}
\usepackage{graphicx}
\usepackage{array}
\usepackage{float}
\usepackage{placeins}
\usepackage{caption}

\usepackage{authblk}

\hypersetup{colorlinks=true,citecolor=blue,linkcolor=blue,urlcolor=blue}
\graphicspath{{figures/}}

\newtheorem{assumption}{Assumption}
\newtheorem{proposition}{Proposition}
\newtheorem{lemma}{Lemma}
\newtheorem{corollary}{Corollary}
\newtheorem{remark}{Remark}

\newcommand{\E}{\mathbb{E}}
\newcommand{\Prob}{\mathbb{P}}
\newcommand{\1}{\mathbf{1}}
\newcommand{\R}{\mathbb{R}}
\newcommand{\Aset}{\mathcal{A}}

\newcommand{\dd}{\,\mathrm{d}}

\title{\bfseries Learning Whom to Trust : Decision-Generated Credibility in Social Learning}
\author[1,2,3]{Gabriel Bontemps\thanks{Corresponding author: gabriel@bontemps.info}}
\author[2,3]{Abhishek Banerjee}
\affil[1]{GREDEG-CNRS, Universit\'e C\^ote d'Azur, Nice, France}
\affil[2]{Blizard Institute, Queen Mary University of London, London, United Kingdom of Great Britain and Northern Ireland (the)}
\affil[3]{Adaptive Decisions Laboratory, Department of Pharmacology, University of Oxford, Oxford, United Kingdom of Great Britain and Northern Ireland (the)}
\date{\today}
\begin{document}
\maketitle

\vspace{10 mm}

\begin{abstract}
\noindent Social interaction can improve collective learning but also amplify early mistakes. We study this tension when the credibility of social information is generated by the sender's own decision process rather than fixed ex ante. Reinforcement-learning agents make binary choices through a drift--diffusion process that jointly determines choice, decision time, and confidence; decision confidence then becomes social credibility by weighting anticipatory influence and retrospective social learning. Under balanced community exposure, the anticipatory field admits an exact quotient representation. Its local Jacobian is a scalar decision-sensitivity term multiplying the community-coupling matrix, which yields a common-mode amplification threshold and an analytical role for cross-community permeability in damping relative community differences. Monte Carlo experiments show the corresponding non-monotone performance pattern: moderate transmission accelerates correction, whereas strong transmission can lock populations into wrong consensus; low permeability instead sustains disagreement. Ablations reveal a dual role for confidence: credibility-sensitive transmission amplifies social error, while confidence-dependent private learning stabilises it. The model yields testable predictions linking sender confidence to receiver behaviour conditional on accuracy.

\end{abstract}
\vspace{20 mm}

\pagebreak
\section{Introduction}

Social interaction can accelerate learning by allowing individuals to use information generated by others. Yet the same process can also propagate early errors, produce persistent disagreement, or coordinate a group on the wrong action. Classical models explain how network structure and updating rules shape information aggregation \citep{DeGroot1974,GolubJackson2010}, while the herding literature shows how social observations can generate inefficient lock-in \citep{Banerjee1992,BikhchandaniHirshleiferWelch1992}. A central question is therefore not whether social influence is beneficial in general, but when the same social environment corrects individual mistakes and when it amplifies them.

This paper studies a behavioural margin that is often treated as exogenous: the \emph{credibility} attached to social information. People do not transmit actions and outcomes as unweighted observations. The force of a social signal depends in part on how certain the sender appears to be. We call this \emph{decision-generated credibility}: decision confidence is produced by the same process that generates the sender's action and becomes socially consequential when receivers use it to weight that signal. Confidence is therefore an intra-agent state, while credibility is the resulting social influence. Formally, baseline network exposure $W_{ij}$ is multiplied by the sender's decision confidence $C_{j,t}$, so effective exposure from $j$ to $i$ varies with the sender's decision process even when the network itself is fixed.

This distinction creates a feedback mechanism. A realised action is generated from learned values through a drift--diffusion process; the same process produces a decision time and a confidence signal. Confidence then affects social learning in two separate ways. In an \emph{anticipatory} channel, neighbours' previous confidence-weighted actions enter current subjective values before choice. In a \emph{retrospective} channel, neighbours' realised outcomes affect subsequent value revision, with the social learning gain depending on their confidence. Confidence also affects private learning, so it can simultaneously increase an agent's influence on others and alter the agent's own responsiveness to prediction errors. The relevant object is therefore a feedback architecture rather than a single social-weight parameter.

The model yields a simple analytical anchor for this feedback. Under balanced community exposure, the anticipatory social field admits an exact community-level representation. Let $B$ denote the row-stochastic community-coupling matrix and let $C_0$ denote process-generated confidence at symmetric indifference. Linearising the representative-community feedback gives the Jacobian
\[
J_\lambda=C_0\frac{\beta a}{\sigma^2}\,\lambda B,
\qquad
\lambda^\star=\frac{\sigma^2}{\beta a C_0}.
\]
Because $B\mathbf 1=\mathbf 1$, the common community mode is locally damped below $\lambda^\star$ and amplified above it. In the symmetric two-community case, the relative or anti-symmetric mode is multiplied by the additional factor $1-2\Gamma$, where $\Gamma$ is cross-community permeability. Thus the same local calculation separates two margins that are distinct in the simulations: $\lambda$ controls whether social perturbations amplify, while permeability controls how strongly community-specific differences survive relative to the common mode. The threshold does not determine the terminal regime of the stochastic learning process, but it identifies the local onset of decision-generated social amplification.

Monte Carlo experiments then establish the global implications. The relationship between social-transmission intensity and collective performance is non-monotone. With little social transmission, learning is slow and groups often remain unresolved over the experimental horizon. At moderate transmission, social information accelerates correction and produces efficient consensus. At stronger transmission, early high-confidence mistakes can be amplified into persistent wrong consensus. Network structure governs the form rather than the existence of this amplification: high cross-community permeability spreads a confident error across communities, whereas low permeability insulates communities and can preserve disagreement. Importantly, the wrong-consensus result does not require a hand-picked initial condition: when agents start from neutral values, stochastic early asymmetries are amplified once the local feedback crosses the analytical threshold.

\paragraph{Contributions.}
The paper makes three contributions. First, it introduces \emph{decision-generated credibility} into a process-based model of social learning. Confidence is generated jointly with choice and decision time rather than assigned externally, and it becomes a time-varying multiplier on social exposure. This links reinforcement learning, drift--diffusion choice and social learning in a way that gives social weight a behavioural origin rather than treating it as a fixed network primitive.

Second, the paper characterises the resulting amplification mechanism. The exact quotient representation isolates the community-level anticipatory field, while the local Jacobian $C_0(\beta a/\sigma^2)\lambda B$ shows how decision parameters and network modes jointly determine amplification. The common mode yields a closed-form threshold and, in the two-community benchmark, permeability attenuates the anti-symmetric mode by the factor $1-2\Gamma$. Simulations then show how these local forces map into efficient consensus, wrong consensus and polarisation in the full stochastic system.

Third, the paper decomposes the mechanism using paired structural counterfactual simulations. At the contested operating point, anticipatory influence, retrospective learning and the decision-time dependence of confidence are required for the observed wrong-consensus regime; credibility-sensitive transmission increases its incidence, while confidence-dependent private learning acts as a brake on error. This decomposition yields direct experimental predictions: receiver behaviour should respond to sender confidence conditional on sender accuracy, and early high-confidence errors should have disproportionately persistent social effects.

\FloatBarrier
\section{Background and positioning}
\label{sec:background}

\paragraph{Social learning, herding and networked learning.}
The primary anchor is the literature on collective learning and opinion
formation in networks \citep{GolubSadler2016}. Classical models of consensus and
naive learning explain when decentralised updating aggregates information
efficiently \citep{DeGroot1974,FriedkinJohnsen1990,GolubJackson2010}, and
continuous-opinion and bounded-confidence models explain when interaction yields
consensus, clustering or persistent disagreement
\citep{DeffuantNeauAmblardWeisbuch2000,HegselmannKrause2002,FlacheMasFelicianiChattoeBrownDeffuantHuetLorenz2017}.
The herding and cascade literature explains how rational or boundedly rational
agents can lock onto incorrect actions when they over-weight others' behaviour
relative to their own signal
\citep{Banerjee1992,BikhchandaniHirshleiferWelch1992,TumpPleskacKurvers2020}.
Cooperative and networked multi-armed bandit models study the same tension
between social acceleration and distortion in an explicit reward-learning
setting, including on block-structured graphs
\citep{LandgrenSrivastavaLeonard2021,XuShanGhaffariWangLiuHajiesmaili2025}, and
reward-like social feedback can drive polarisation directly even on fixed
networks \citep{BanischOlbrich2019,HornBanischBatzdorferReitenbachSartoriSchwabeMas2026}.
Our model contributes a mechanism to this tradition: the weight an agent places
on a neighbour is not fixed but varies with decision-generated credibility, because
the relevant confidence signal is produced by that neighbour's own decision process.

\paragraph{What the model changes relative to adjacent traditions.}
Table~\ref{tab:comparison} positions the model. Relative to DeGroot and
bounded-confidence opinion models, agents do not update opinions by weighted
averaging alone: they learn from stochastic rewards, choose through a process
model, and transmit confidence-marked actions and outcomes. Relative to herding
and cascade models, the credibility attached to social information is not
exogenously given but generated by the neighbour's own decision process.
Relative to cooperative bandit models, the model separates anticipatory
influence on choice from retrospective influence on value revision. Relative to
polarisation ABMs, modular divergence arises from the interaction between
confidence-weighted transmission and reward learning rather than from fixed
homophily or ideological repulsion alone; and relative to
reinforcement-learning drift--diffusion (RLDDM) accounts of individual choice,
the novelty is the social transmission of decision-generated credibility on a network.

\begin{table}[H]\centering\small
\caption{Positioning relative to adjacent model families.}
\label{tab:comparison}
\begin{tabular}{p{3.1cm}p{3.2cm}p{2.9cm}p{3.4cm}}
\toprule
Tradition & Typical mechanism & Exogenous element & What this paper changes \\
\midrule
DeGroot / consensus & weighted opinion averaging & weights, signals & confidence generated by the choice process \\
Herding / cascades & action inference & reliability of social signal & credibility is decision-generated \\
Cooperative bandits & reward learning across agents & communication weight & two distinct social channels \\
Polarisation ABMs & homophily / social feedback & confidence usually absent & confidence-driven amplification \\
RLDDM (individual) & value-based diffusion choice & no social network & social transmission of credibility \\
\bottomrule
\end{tabular}
\end{table}

\paragraph{Agent-based social simulation and documentation.}
The model is a mechanism-oriented agent-based simulation: aggregate regimes emerge from explicit individual learning and decision rules rather than being imposed directly \citep{StoneKunasekaranPoulosMacIntyreHeslop2026}. We document the implementation using the ODD conventions and the ODD+D extension for models with human decision-making \citep{GrimmBergerDeAngelisPolhillGiskeRailsback2010,MullerBohmeFrankDresslerGroeneveldKlassertMartinSchluterSchulzeWeiseSchwarz2013}. The second ODD update explicitly recommends a concise journal-level summary accompanied by a complete description for replication, and adds greater emphasis on model rationale, evaluation and links to code \citep{GrimmRailsbackVincenotEtAl2020}. We follow that division: the main text states the behavioural mechanism and analytical results, while Appendix~\ref{app:odd} gives the full ODD+D description and design rationale and the subsequent appendices document verification, parameters and robustness. Synchronous trial timing and simultaneous post-outcome updating are fixed modelling choices, and modular structure is represented through interpretable community-coupling parameters rather than an unrestricted graph family \citep{Axtell2000,LeeWilkinson2019,Peixoto2014}.

\paragraph{Computational models of social cognition and reinforcement learning.}
Contemporary computational accounts treat social learning with the same
primitives as non-social value learning --- value, prediction error,
uncertainty and learning rate
\citep{RuffFehr2014,LockwoodKleinFlugge2021}. This licenses our decision to let
social information enter the same latent-value machinery as private reward, and
to allow learning rates that depend on confidence rather than being globally
fixed \citep{WangVeismannBanerjeePleger2023}. In this reading the retrospective
channel is an \emph{other-referenced} prediction error, the social counterpart of
the self-referenced reward prediction error, consistent with evidence that social
and non-social learning share prediction-error machinery in the brain
\citep{JoinerPivaTurrinChang2017}.

\paragraph{Process-based choice and confidence.}
Drift--diffusion and reinforcement-learning-diffusion models make choice a
within-trial accumulation process and render decision time informative
\citep{Ratcliff1978,RatcliffSmithBrownMcKoon2016,PedersenFrankBiele2017,FontanesiGluthSpektorRieskamp2019}.
The diffusion limit is statistically optimal for two-alternative choice
\citep{BogaczBrownMoehlisHolmesCohen2006} and admits an economic optimal-stopping
foundation in which the decision time is itself chosen
\citep{FudenbergStrackStrzalecki2018}. Confidence can be read from the same
accumulation process \citep{KianiShadlen2009,PleskacBusemeyer2010,Drugowitsch2019},
can regulate learning, and may remain miscalibrated
\citep{SalemGarciaPalminteriLebreton2023}; communicated confidence can even
improve joint decisions \citep{BahramiOlsenLathamRoepstorffReesFrith2010}, which
is exactly the channel that the present model makes double-edged. We use these
results as modelling restrictions: confidence is generated by the decision process, depends on evidence
strength and decision time, regulates learning, and is \emph{not} equated with
accuracy.

\paragraph{Why confidence is socially transmitted.}
Confidence is useful socially because it is a compact signal about the strength of the decision process that generated an action. A receiver can condition on that signal without reconstructing the sender's latent computation. This interpretation is consistent with computational accounts in which social and non-social learning use related value and prediction-error primitives \citep{JoinerPivaTurrinChang2017,LockwoodKleinFlugge2021}. We therefore use confidence as a behaviourally interpretable input into social weighting, without equating it with objective accuracy. The neuroscience literature motivates the process variables; the claims of the present paper are about behavioural and computational mechanisms.

\FloatBarrier
\section{The model}
\label{sec:model}

\begin{figure}[H]
\centering
\includegraphics[width=\textwidth]{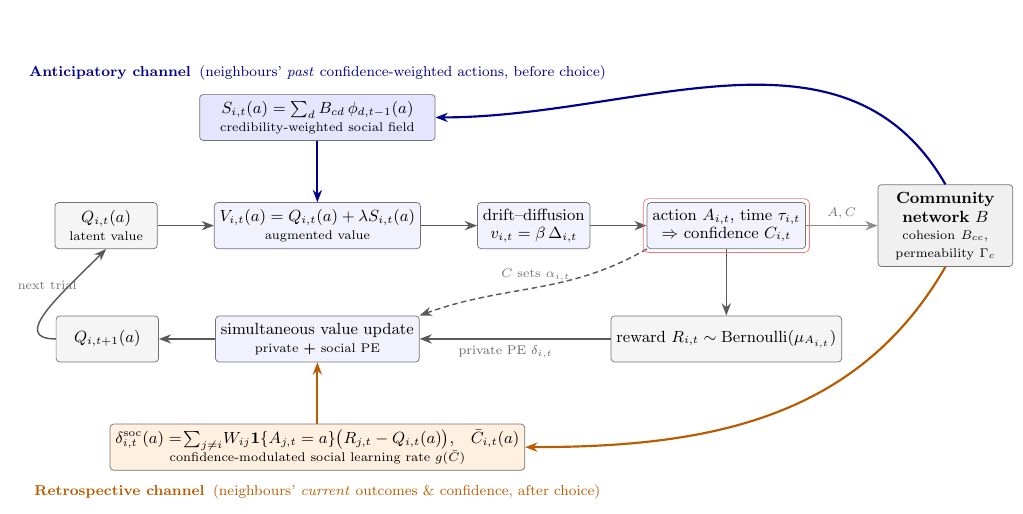}
\caption{Mechanism. Within a trial, latent values $Q_{i,t}$ are augmented by the
anticipatory credibility field $S_{i,t}$ (top, blue) to form subjective values
$V_{i,t}$; a drift--diffusion process turns the value contrast into an action
$A_{i,t}$, a decision time $\tau_{i,t}$ and process-generated confidence $C_{i,t}$
(red box). After the reward is observed, values are revised by a simultaneous
update combining the private prediction error with the retrospective social
prediction error (bottom, orange), whose strength is modulated by neighbours'
confidence. Confidence is the shared link between the two channels. The
community coupling matrix $B$ governs how credibility circulates.}
\label{fig:mechanism}
\end{figure}

Following the updated ODD guidance, this section provides a concise mathematical specification of the model, while Appendix~\ref{app:odd} supplies the complete ODD+D description, including the rationale for the decision submodels and model evaluation \citep{GrimmRailsbackVincenotEtAl2020,MullerBohmeFrankDresslerGroeneveldKlassertMartinSchluterSchulzeWeiseSchwarz2013}. Figure~\ref{fig:mechanism} summarises the within-trial mechanism and the two social channels.

\FloatBarrier
\subsection{Purpose and entities}
The model's purpose is to explain when social transmission under decision-generated credibility
damps early errors and when it amplifies them. The
entities are \emph{agents} and \emph{communities}. Agents choose repeatedly
between two arms of a stochastic bandit, learn from reward, and influence one
another through a weighted communication network. Communities are an exogenous
partition of agents into modules.

\FloatBarrier
\subsection{State variables and environment}
Time is discrete at the trial level, $t\in\mathbb N_0$. Agents are
$V=\{1,\dots,N\}$, partitioned into communities $V=\bigsqcup_{c=1}^M V_c$ with
$N_c=|V_c|$; community membership is $g(i)$. The action set is
$\Aset=\{1,2\}$. The weighted communication matrix
$W=(W_{ij})\in\R_+^{N\times N}$ is row-stochastic, $\sum_j W_{ij}=1$. For each
agent $i$, trial $t$ and arm $a$ we track the latent value $Q_{i,t}(a)$, the
anticipatory social signal $S_{i,t}(a)$, the augmented value $V_{i,t}(a)$, the
value contrast $\Delta_{i,t}$, the drift $v_{i,t}$, the action $A_{i,t}$, the
reward $R_{i,t}\in\{0,1\}$, the decision time $\tau_{i,t}$ and the confidence
$C_{i,t}\in(0,1)$. Arm $a$ pays Bernoulli reward with mean $\mu_a\in(0,1)$;
$a^\star\in\arg\max_a\mu_a$, $\mu^\star=\max_a\mu_a$ and $\Delta\mu=\mu_1-\mu_2$.

\FloatBarrier
\subsection{Process overview and scheduling}
Each trial uses a synchronous schedule with a simultaneous post-outcome update:
(1) construct the anticipatory social signal from the previous period; (2) form
augmented values; (3) generate the drift--diffusion choice and decision time;
(4) read out confidence; (5) draw rewards; (6) compute all private and social
prediction errors from the \emph{pre-update} values; (7) apply private and
social updates simultaneously. Step~6 prevents order artefacts between private
and social revision.

\paragraph{Information structure.}
Before choosing, agents observe neighbours' previous actions and confidence
signals. After outcomes are realised, they observe neighbours' actions, realised
outcomes and confidence for the retrospective channel. Confidence is interpreted
as communicated or behaviourally inferred subjective reliability, not as
objective accuracy.

\paragraph{Confidence and credibility.}
We distinguish the sender's \emph{confidence} from the receiver's resulting social exposure. $C_{j,t}$ is an intra-agent state generated by agent $j$'s decision process. We refer to the mapping from this decision confidence into receiver-specific social exposure as \emph{decision-generated credibility}. In the anticipatory channel, the effective credibility-weighted exposure from $j$ to $i$ is $W_{ij}C_{j,t}$: $W_{ij}$ is the baseline network weight and $C_{j,t}$ is a time-varying multiplier generated by the sender's decision process. Equation~\eqref{eq:Si} is the operative definition used in the model and code. Normalised confidence weights that appear in schematic illustrations should therefore be read as visual summaries of relative credibility, not as a replacement for $W_{ij}C_{j,t}$ in the law of motion.

\FloatBarrier
\subsection{Submodels}

\paragraph{Anticipatory social signal.}
Social information available before the current choice is
\begin{equation}
S_{i,t}(a):=\sum_{j=1}^N W_{ij}\,\1\{A_{j,t-1}=a\}\,C_{j,t-1},
\qquad S_{i,0}(a)=0,
\label{eq:Si}
\end{equation}
the baseline network exposure to neighbours who chose $a$, multiplied by their
process-generated confidence. Subjective values are socially augmented,
\begin{equation}
V_{i,t}(a):=Q_{i,t}(a)+\lambda\,S_{i,t}(a),
\qquad
\Delta_{i,t}:=V_{i,t}(1)-V_{i,t}(2),
\label{eq:V}
\end{equation}
where $\lambda\ge 0$ is the anticipatory weight.

\paragraph{Process-based choice and decision time.}
Conditional on $\Delta_{i,t}$, the within-trial decision variable is a Wiener
process with drift,
\begin{equation}
\dd X_{i,t}(s)=v_{i,t}\dd s+\sigma\dd B_{i,t}(s),\quad X_{i,t}(0)=0,\quad
v_{i,t}=\beta\,\Delta_{i,t},
\label{eq:ddm}
\end{equation}
absorbed at symmetric boundaries $\pm a$; the action is $A_{i,t}=1$ if the upper
boundary is hit first and $A_{i,t}=2$ otherwise, with first-passage time
$\tau_{i,t}$. For a constant drift $v$, the standard symmetric two-boundary
Wiener first-passage identities are
\begin{equation}
\Prob(A=1\mid v)=\frac{1}{1+\exp(-2va/\sigma^2)},
\label{eq:choiceprob}
\end{equation}
and
\begin{equation*}
\bar\tau(v):=\E[\tau\mid v]=
\begin{cases}
\displaystyle \frac{a}{v}\tanh\!\left(\frac{av}{\sigma^2}\right), & v\neq0,\\[7pt]
\displaystyle \frac{a^2}{\sigma^2}, & v=0.
\end{cases}
\tag{4a}\label{eq:meanfpt}
\end{equation*}
These identities follow from the scale-function/optional-stopping argument for
the hitting probability and Dynkin's equation for the mean exit time; they are
standard in drift--diffusion and Brownian first-passage theory
\citep{Ratcliff1978,BogaczBrownMoehlisHolmesCohen2006,KaratzasShreve1991}.
Appendix~\ref{app:proofs} gives the derivation used here. It also records four
properties needed below: $\bar\tau(v)$ is even in $v$, strictly positive,
bounded above by $a^2/\sigma^2$, and strictly decreasing in $|v|$ away from
zero. Thus stronger value evidence produces, in expectation, faster decisions.

The implementation uses the exact choice probability \eqref{eq:choiceprob} but
approximates the joint choice--time law: reaction times are drawn from a
positive, mean-matched inverse-Gaussian whose mean equals \eqref{eq:meanfpt} and
whose squared coefficient of variation is a fixed dispersion, rather than from
the exact two-boundary first-passage density. Appendix~\ref{app:tests} validates
the choice probability and mean reaction time against a direct
Euler--Maruyama simulation of \eqref{eq:ddm}, and Appendix~\ref{app:robust}
shows that regime classification is insensitive to this approximation.

\paragraph{Process-generated confidence.}
Confidence is read out from the same process and is bounded in $(0,1)$:
\begin{equation}
C_{i,t}=\Lambda\!\left(\kappa_1\frac{|v_{i,t}|}{a}
-\kappa_2\log\!\Big(1+\frac{\tau_{i,t}}{\tau_0}\Big)\right),
\qquad \Lambda(x)=\frac{1}{1+e^{-x}},
\label{eq:confidence}
\end{equation}
with $\kappa_1,\kappa_2,\tau_0>0$. Confidence rises with evidence strength and
falls with decision time. It is a subjective reliability weight and is
\emph{not} assumed to equal correctness; this separation is what allows
high-confidence error. An alternative balance-of-evidence map, in which
confidence equals the DDM probability of the chosen boundary, is used in the
ablations.

To make the process interpretation explicit, define the mean-ridge confidence
obtained by evaluating \eqref{eq:confidence} at the process's own mean decision
time,
\begin{equation*}
\bar C(\Delta):=\Lambda\!\left(
\kappa_1\frac{\beta|\Delta|}{a}
-\kappa_2\log\!\left(1+\frac{\bar\tau(\beta\Delta)}{\tau_0}\right)
\right).
\tag{5a}\label{eq:confridge}
\end{equation*}

\begin{lemma}[Properties of process-generated confidence]
\label{lem:confidence}
For $\kappa_1,\kappa_2,\tau_0>0$: (i) $0<C<1$; holding decision time fixed,
confidence is strictly increasing in evidence magnitude $|v|$, and holding
$|v|$ fixed it is strictly decreasing in $\tau$; (ii) $\bar C(\Delta)$ is even
and strictly increasing in $|\Delta|$ away from indifference; and (iii)
\begin{equation*}
C_0:=\bar C(0)
=\frac{1}{1+\left(1+a^2/(\sigma^2\tau_0)\right)^{\kappa_2}}
\in(0,1/2).
\tag{5b}\label{eq:C0}
\end{equation*}
Moreover $\bar C(\Delta)=C_0+O(|\Delta|)$ as $\Delta\to0$. Because of the
absolute-value term, the confidence map generally has a kink at indifference;
the later amplification result therefore uses Lipschitz continuity, not a
Taylor derivative of confidence at zero.
\end{lemma}

The choice probability satisfies $p(-\Delta)=1-p(\Delta)$ while
$\bar C(-\Delta)=\bar C(\Delta)$. Hence opposite value contrasts of equal
magnitude carry the same process-generated confidence even though they favour
opposite actions. When one arm is objectively optimal, a sufficiently negative
subjective contrast can therefore be both wrong-directed and highly confident.
Appendix~\ref{app:proofs} proves Lemma~\ref{lem:confidence} and makes this
``high-confidence error'' symmetry explicit.

\begin{figure}[H]
\centering
\includegraphics[width=\textwidth]{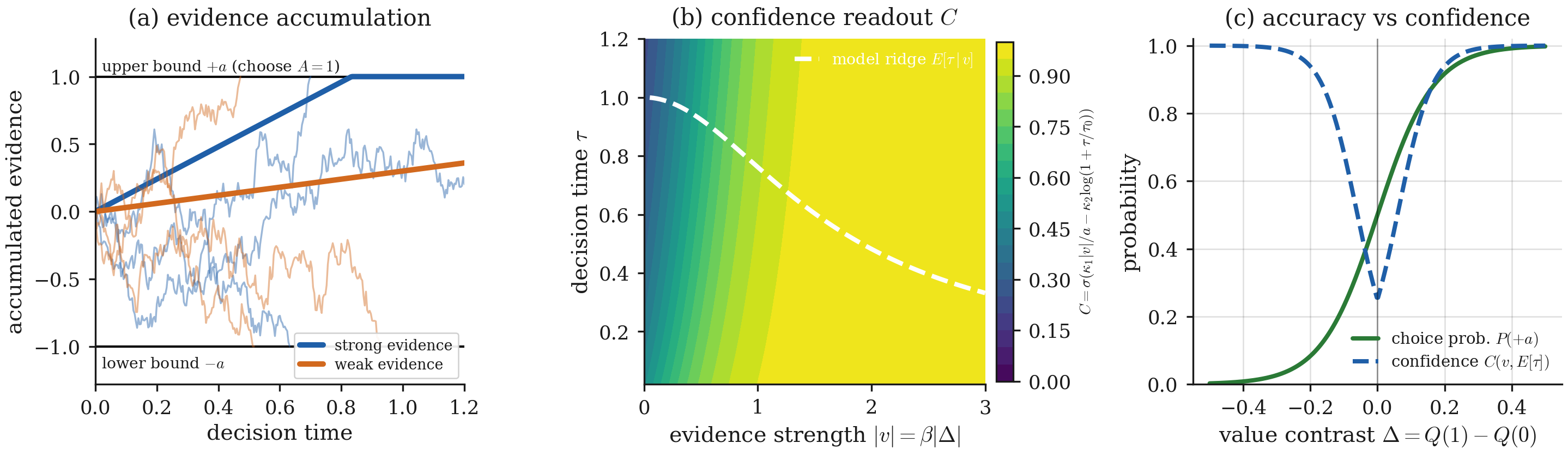}
\caption{Confidence as a decision-process readout (all curves are the
exact model maps, not fits). \textbf{(a)}~Evidence accumulates to a boundary
$\pm a$ at rate $v=\beta\Delta$; strong evidence yields fast, confident choices,
weak evidence slow, uncertain ones. \textbf{(b)}~Confidence
$C=\Lambda(\kappa_1|v|/a-\kappa_2\log(1+\tau/\tau_0))$ rises with evidence
strength and falls with decision time; the dashed ridge is the process's own
mean decision time $\E[\tau\mid v]$. \textbf{(c)}~Choice accuracy $P(+a)$ is
monotone in the value contrast, but confidence is \emph{U-shaped} in $\Delta$ ---
it tracks evidence \emph{magnitude}, not correctness --- so a strongly held wrong
belief ($\Delta<0$, large) is transmitted with high credibility. This decoupling of confidence from accuracy is the decision-process basis for high-confidence error in the model.}
\label{fig:decisionconf}
\end{figure}

\paragraph{Private value revision.}
The private prediction error is $\delta_{i,t}=R_{i,t}-Q_{i,t}(A_{i,t})$, with a
confidence-modulated learning rate
\begin{equation}
\alpha_{i,t}=
\begin{cases}
\alpha^{\min}+(\alpha^{\max}-\alpha^{\min})\,C_{i,t}, & \delta_{i,t}<0,\\[2pt]
\alpha^{\max}-(\alpha^{\max}-\alpha^{\min})\,C_{i,t}, & \delta_{i,t}\ge 0,
\end{cases}
\label{eq:privatelr}
\end{equation}
so higher confidence increases the learning rate applied to negative prediction
errors and decreases the rate applied to non-negative prediction errors. Both
branches lie in $[\alpha^{\min},\alpha^{\max}]$. Writing them as
$\alpha^{-}(C)$ and $\alpha^{+}(C)$ respectively,
\[
\alpha^{-}(C)-\alpha^{+}(C)
=(\alpha^{\max}-\alpha^{\min})(2C-1),
\]
so the two branches cross exactly at $C=1/2$: bad news receives the larger gain
if and only if confidence exceeds one half. Appendix~\ref{app:proofs} records
the elementary bounds.

\begin{figure}[H]
\centering
\includegraphics[width=0.82\textwidth]{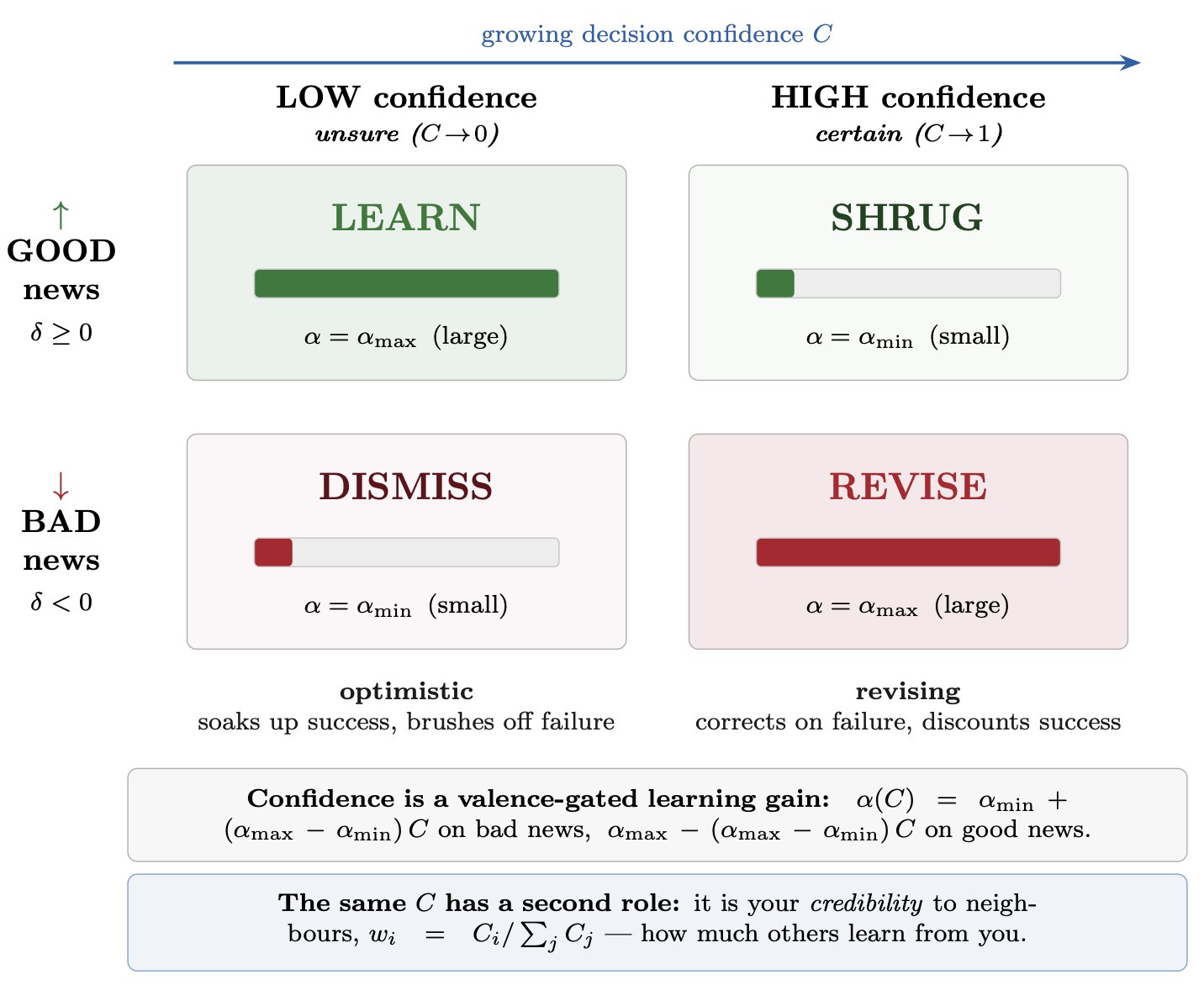}
\caption{Confidence is a valence-gated learning gain. Reading the $2\times2$
matrix: the confidence-modulated learning rate \eqref{eq:privatelr} is largest
($\alpha_{\max}$) for \emph{good} news under \emph{low} confidence and for
\emph{bad} news under \emph{high} confidence, and smallest ($\alpha_{\min}$) on
the other diagonal. A low-confidence agent is therefore optimistic (soaks up
success, brushes off failure); a high-confidence agent is revising (corrects on
failure, discounts success) --- a confidence-gated analogue of asymmetric
reward-prediction-error learning. The same confidence \(C\) plays a second role in social transmission. The normalised quantity \(w_i=C_i/\sum_j C_j\) shown in the schematic summarises relative confidence within the local signal pool; the operative anticipatory exposure in the model remains \(W_{ij}C_{j,t}\) as defined in Eq.~\eqref{eq:Si}. Confidence therefore acts both as an intra-agent learning gain and as a decision-generated multiplier on baseline social exposure. The exact curves are plotted in Appendix~\ref{app:plasticity} (Figure~\ref{fig:plasticityquant}).}
\label{fig:plasticity}
\end{figure}

\paragraph{Retrospective social learning.}
The retrospective channel uses neighbours' realised outcomes. For each arm,
\begin{equation}
\delta^{\mathrm{soc}}_{i,t}(a):=\sum_{j\neq i}W_{ij}\,\1\{A_{j,t}=a\}\bigl(R_{j,t}-Q_{i,t}(a)\bigr),
\quad
\bar C_{i,t}(a):=\frac{\sum_{j\neq i}W_{ij}\,\1\{A_{j,t}=a\}\,C_{j,t}}
{\sum_{j\neq i}W_{ij}\,\1\{A_{j,t}=a\}+\varepsilon}.
\label{eq:socpe}
\end{equation}
We adopt the convention that \emph{confidence modulates the social learning
rate} (rather than re-weighting the social prediction error itself): the social
learning rate is $g(\bar C_{i,t}(a))=\gamma\,\bar C_{i,t}(a)^{\omega}$ with
$\gamma>0,\omega>0$, $g(0)=0$. The simultaneous law of motion, with all
prediction errors formed from $Q_{i,t}$, is
\begin{equation}
Q_{i,t+1}(a)=\Pi_{[0,1]}\!\Big(
Q_{i,t}(a)+\1\{A_{i,t}=a\}\,\alpha_{i,t}\,\delta_{i,t}
+\eta\,g\!\big(\bar C_{i,t}(a)\big)\,\delta^{\mathrm{soc}}_{i,t}(a)\Big),
\label{eq:update}
\end{equation}
where $\eta\ge0$ is the retrospective weight and $\Pi_{[0,1]}$ projects onto the
reward support, keeping values interpretable as Bernoulli-mean estimates. Since
$0<C_{j,t}<1$ and $\varepsilon>0$, the regularised neighbour average satisfies
$0\le \bar C_{i,t}(a)<1$ even when no neighbour chooses arm $a$. Therefore
$0\le g(\bar C_{i,t}(a))<\gamma$ for $\omega>0$, and the projection makes
$[0,1]^{2N}$ forward-invariant for the latent values. These bounds are proved in
Appendix~\ref{app:proofs}.

\paragraph{Self-weighting convention.}
We exclude an agent's own current outcome from the retrospective channel
(the sum in \eqref{eq:socpe} is over $j\neq i$), since that outcome is already
captured by the private prediction error. We \emph{include} the agent's own
previous confidence-weighted action in the anticipatory signal \eqref{eq:Si}
(so $W_{ii}\ge0$), interpreting it as choice inertia. This convention makes the
quotient field of Section~\ref{sec:quotient} exact; the verification suite tests
both inclusions (Appendix~\ref{app:tests}).

\FloatBarrier
\section{Community aggregation and local amplification}
\label{sec:quotient}

We first study the smallest non-trivial modular society and then generalise.
For agent $i\in V_c$ and community $d$, define the block exposure
$b_{i\to d}:=\sum_{j\in V_d}W_{ij}$.

\begin{assumption}[Balanced block weights]
\label{ass:balanced}
The population is partitioned into $M$ non-empty communities
$V_1,\ldots,V_M$, with $N_d:=|V_d|>0$. There exists a non-negative
row-stochastic matrix $B=(B_{cd})_{c,d=1}^M$ such that
\[
W_{ij}=\frac{B_{cd}}{N_d}
\qquad\forall i\in V_c,\;j\in V_d.
\]
\end{assumption}

The central mesoscopic object is the confidence-weighted action mass
\begin{equation}
\phi_{c,t}(a):=\frac1{N_c}\sum_{i\in V_c}\1\{A_{i,t}=a\}\,C_{i,t},
\label{eq:phi}
\end{equation}
which combines the fraction choosing $a$ with the average confidence carried by
those choices. We also record the action frequency
$m_{c,t}(a)=N_c^{-1}\sum_{i\in V_c}\1\{A_{i,t}=a\}$ and the mean value
$\bar Q_{c,t}(a)$.

\begin{proposition}[Exact quotient identity]
\label{prop:quotient}
Under Assumption~\ref{ass:balanced}, let $f_j$ be any scalar agent-level
statistic and let $\bar f_d=N_d^{-1}\sum_{j\in V_d}f_j$. Then, for every
$i\in V_c$,
\begin{equation*}
\sum_{j=1}^N W_{ij}f_j=\sum_{d=1}^M B_{cd}\,\bar f_d.
\tag{9a}\label{eq:quotientgeneral}
\end{equation*}
In particular,
\begin{equation}
S_{i,t}(a)=\sum_{d=1}^M B_{cd}\,\phi_{d,t-1}(a),
\label{eq:quotientfield}
\end{equation}
so the anticipatory social field is identical for all agents in the same
community. Assumption~\ref{ass:balanced} also implies
$\sum_jW_{ij}=1$ and $b_{i\to d}=B_{cd}$.
\end{proposition}

\begin{corollary}[Finite-population self-exclusion]
\label{cor:selfexclude}
For $i\in V_c$,
\[
\sum_{j\neq i}W_{ij}f_j
=\sum_{d=1}^M B_{cd}\bar f_d-\frac{B_{cc}}{N_c}f_i.
\]
Hence, if $|f_i|\le1$, replacing a self-excluded weighted sum by its quotient
aggregate introduces an error of at most $B_{cc}/N_c$.
\end{corollary}

Equation~\eqref{eq:quotientfield} is the precise sense in which the community
network is a quotient of the microscopic network for the anticipatory channel.
The diagonal $B_{cc}$ measures within-community cohesion and
$\Gamma_c:=1-B_{cc}=\sum_{d\neq c}B_{cd}$ measures cross-community
permeability. Proposition~\ref{prop:quotient} is exact; the
$O(N_c^{-1})$ correction that appears when the retrospective channel excludes
self is quantified by Corollary~\ref{cor:selfexclude}. Under balanced weights,
these identities also reduce the computational cost of the social aggregates
from $O(N^2)$ to $O(N)$ per trial.

Define the confidence-weighted community gap and the private-value gap by
\begin{equation}
\psi_{c,t}:=\phi_{c,t}(1)-\phi_{c,t}(2),
\qquad
\bar\Delta^Q_{c,t}:=\bar Q_{c,t}(1)-\bar Q_{c,t}(2),
\label{eq:gaps}
\end{equation}
and collect them in vectors
$\bm\psi_t=(\psi_{1,t},\ldots,\psi_{M,t})^\top$ and
$\bar{\bm\Delta}^Q_t=(\bar\Delta^Q_{1,t},\ldots,\bar\Delta^Q_{M,t})^\top$.
The quotient identity gives the exact social component of the community
contrast,
\begin{equation}
\bar{\bm\Delta}_t
=\bar{\bm\Delta}^Q_t+\lambda B\bm\psi_{t-1}.
\label{eq:vectorgap}
\end{equation}
The private term is disciplined by realised rewards; the second term feeds the
existing confidence-weighted mass back into the next decision.

\begin{remark}[Exact field, approximate value dynamics]
\label{rem:exact}
Equations~\eqref{eq:quotientgeneral}--\eqref{eq:vectorgap} are exact under
Assumption~\ref{ass:balanced}. The value dynamics remain stochastic at the
agent level because rewards and diffusion noise are individual. A deterministic
representative-community recursion is therefore an analytical approximation to
the full value process, not an exact large-population law. We quantify its
accuracy in Section~\ref{sec:results-quotient}.
\end{remark}

\FloatBarrier
\subsection{Local network-mode amplification}
\label{sec:localamp}

The exact quotient makes it possible to isolate the anticipatory feedback
without replacing network structure by a scalar ``effective coupling.'' In the
deterministic representative-community reduction, the arm-1 choice probability
at contrast $\Delta$ is
\[
p(\Delta)=\Lambda\!\left(\frac{2\beta a}{\sigma^2}\Delta\right),
\]
so its action-frequency gap is
\begin{equation}
h(\Delta):=2p(\Delta)-1
=\tanh\!\left(\frac{\beta a}{\sigma^2}\Delta\right)
=\kappa\Delta+O(\Delta^3),
\qquad \kappa:=\frac{\beta a}{\sigma^2}.
\label{eq:actiongaplin}
\end{equation}
Lemma~\ref{lem:confidence} gives
$\bar C(\Delta)=C_0+O(|\Delta|)$. Therefore the representative
confidence-weighted mass gap satisfies
\begin{equation}
\bar C(\Delta)h(\Delta)
=C_0\kappa\Delta+O(\Delta^2).
\label{eq:massgaplin}
\end{equation}
The $O(\Delta^2)$ remainder is the key point: the kink in confidence at
indifference does not alter the first-order amplification multiplier.

\begin{proposition}[Local network-mode amplification]
\label{prop:amplification}
Linearise the deterministic representative-community recursion around symmetric
indifference. Then
\begin{equation}
\bm\psi_t
=C_0\kappa\left(\bar{\bm\Delta}^Q_t+\lambda B\bm\psi_{t-1}\right)
+\bm r_t,
\qquad
\|\bm r_t\|_\infty=O(\|\bar{\bm\Delta}_t\|_\infty^2),
\label{eq:vectorlin}
\end{equation}
and, holding the local private-value gap fixed, the Jacobian of the
anticipatory loop is
\begin{equation}
J_\lambda=C_0\kappa\lambda B.
\label{eq:jacobian}
\end{equation}
Because $B$ is non-negative and row-stochastic, $\|B\|_\infty=1$ and
$B\mathbf 1=\mathbf 1$ \citep[see, e.g.,][]{HornJohnson2013}. Hence:
\begin{enumerate}[label=(\roman*),leftmargin=2em]
\item if $C_0\kappa\lambda<1$, the linearised anticipatory loop is a contraction
in the $\ell_\infty$ norm and every network perturbation is locally damped;
\item if $C_0\kappa\lambda>1$, the common mode $\mathbf 1$ is locally amplified;
\item for any eigenmode $Bq=\xi q$, the one-step multiplier is
$\rho_\xi=C_0\kappa\lambda\xi$.
\end{enumerate}
The common-mode threshold is therefore
\begin{equation}
\lambda^\star
=\frac{1}{C_0\kappa}
=\frac{\sigma^2}{\beta a C_0}.
\label{eq:lambdastar}
\end{equation}
The true state remains bounded, $|\psi_{c,t}|\le1$; ``amplification'' therefore
means local escape from the neighbourhood of symmetric indifference, not
unbounded growth of the actual agent system.
\end{proposition}

\begin{corollary}[Two-community permeability modes]
\label{cor:twomodes}
For two symmetric communities with
\begin{equation}
B(\Gamma)=
\begin{pmatrix}
1-\Gamma & \Gamma\\
\Gamma & 1-\Gamma
\end{pmatrix},
\qquad 0\le\Gamma\le\tfrac12,
\label{eq:Bgamma}
\end{equation}
the common mode $q_+=(1,1)^\top$ and anti-symmetric mode
$q_-=(1,-1)^\top$ have multipliers
\begin{equation}
\rho_+=C_0\kappa\lambda,
\qquad
\rho_-=C_0\kappa\lambda(1-2\Gamma).
\label{eq:twomultipliers}
\end{equation}
Thus permeability does not change the common-mode threshold
\eqref{eq:lambdastar}, but on $[0,1/2]$ it monotonically attenuates relative
community differences. The corollary does not by itself prove polarisation or
wrong consensus in the nonlinear stochastic system; it identifies the local
network mode through which permeability changes the form of amplification.
\end{corollary}

For the default decision parameters, $a=\sigma=1$, $\tau_0=0.5$ and
$\kappa_2=1$, Eq.~\eqref{eq:C0} gives $C_0=1/4$ exactly. With $\beta=6$,
Eq.~\eqref{eq:lambdastar} therefore gives
$\lambda^\star=2/3\simeq0.67$, close to the transition in the simulated phase
diagram (Figure~\ref{fig:phase}). A neutral-initialisation test without an
imposed early lead is reported in Appendix~\ref{app:earlylead}.

Holding $C_0$ fixed, the threshold has the transparent partial comparative
statics
\begin{equation}
\frac{\partial\lambda^\star}{\partial\beta}=-\frac{\lambda^\star}{\beta}<0,
\quad
\frac{\partial\lambda^\star}{\partial a}=-\frac{\lambda^\star}{a}<0,
\quad
\frac{\partial\lambda^\star}{\partial\sigma}=\frac{2\lambda^\star}{\sigma}>0,
\quad
\frac{\partial\lambda^\star}{\partial C_0}=-\frac{\lambda^\star}{C_0}<0.
\label{eq:partialcs}
\end{equation}
Because $C_0$ is itself generated by the decision process, these are partial
rather than total effects. Substituting Eq.~\eqref{eq:C0} yields the fully
reduced local threshold
\begin{equation}
\lambda^\star
=\frac{\sigma^2}{\beta a}
\left[1+\left(1+\frac{a^2}{\sigma^2\tau_0}\right)^{\kappa_2}\right].
\label{eq:lambdastarstructural}
\end{equation}
Two implications follow directly. First, $\kappa_1$ does not enter the local
onset of amplification because evidence magnitude vanishes at indifference;
$\kappa_1$ instead governs how quickly credibility changes once the system
moves away from the local neighbourhood. Second, stronger decision-time
penalisation $\kappa_2$ raises $\lambda^\star$, while a larger time scale
$\tau_0$ lowers it. The total effects of $a$ and $\sigma$ combine their direct
effects on choice sensitivity with their indirect effects through $C_0$ and
need not coincide with the partial signs in Eq.~\eqref{eq:partialcs}. Full
derivations are in Appendix~\ref{app:proofs}.

\FloatBarrier
\section{Simulation design}
\label{sec:design}

\paragraph{Observables and regimes.}
We measure group regret
$\mathcal R_T=\sum_{i,t\le T}(\mu^\star-\mu_{A_{i,t}})$, time to consensus, the
polarisation index $\Pi_t=|m_{1,t}(1)-m_{2,t}(1)|$, correction lag after an
early wrong lead, and, for the quotient study, the micro--meso discrepancy.
Terminal outcomes are classified into four operational regimes using a $0.9$
mass threshold: \emph{efficient consensus} (every community places
$\ge0.9$ on $a^\star$), \emph{wrong consensus} (every community places $\ge0.9$
on the suboptimal arm), \emph{polarisation} (communities lock on different
arms), and \emph{unresolved} otherwise.

\paragraph{Monte Carlo and ablations.}
Each configuration is replicated over independent reward and diffusion draws;
reported regime probabilities are frequencies over replications. The design
varies the reward gap, the anticipatory weight $\lambda$, the retrospective
weight $\eta$, cohesion and permeability. Five decisive ablations isolate the
mechanism: $\lambda=0$ (no anticipatory channel), $\eta=0$ (no retrospective
channel), constant learning rates, no confidence weighting of transmission, and
the alternative balance-of-evidence confidence map.

\FloatBarrier
\subsection{Structural counterfactuals and Monte Carlo inference}
\label{sec:identification}
Because the stochastic data-generating process is fully specified, disabling one model component while holding the remaining structure fixed defines a transparent model-internal counterfactual. We use these counterfactuals to decompose the mechanism and pair them with Monte Carlo uncertainty rather than treating differences across unrelated simulations as evidence about component effects.

\paragraph{Estimands.} We report three objects: (i) regime probabilities, the frequencies of terminal regimes across independent replications at a fixed configuration; (ii) mean group regret and correction lag; and (iii) \emph{structural component contrasts}. Writing $c=0$ for the variant in which component $c$ is disabled, define
\[
\tau_c(\theta):=\E\big[Y\mid \theta,\text{full}\big]-\E\big[Y\mid \theta,c=0\big].
\]
The contrast is internal to the specified model: it asks how the outcome changes when one structural component is switched off while all other equations and parameters are held fixed. It should therefore be interpreted as a mechanism decomposition of the model rather than as an econometric treatment effect identified from observational data.

\paragraph{Paired simulation under common random numbers.} We estimate $\tau_c$ by simulating the full model and the corresponding ablation under \emph{common random numbers}. Each replication pre-generates the same exogenous random environment---a uniform reward draw and a decision-noise draw for every agent and trial, indexed by the same seed---and each structural variant then maps those draws through its own endogenous action history. The paired difference
\[
D_r=Y_r(\text{full})-Y_r(c=0)
\]
therefore compares the two structural variants under the same exogenous simulation shocks. This coupling does not create identification; the counterfactual is defined by the model itself. Its role is to make the component comparison ceteris paribus within a replication and to reduce Monte Carlo variance relative to independently simulated variants.

\paragraph{Uncertainty reporting.} Regime probabilities are empirical
frequencies over independent Monte Carlo replications; we report Wilson $95\%$
intervals for binary regime probabilities and Monte Carlo standard errors with
nonparametric percentile-bootstrap $95\%$ intervals (resampling replications) for
regret and correction lag. Counterfactual contrasts $\widehat\tau_c$ carry
paired-bootstrap $95\%$ intervals (resampling replication indices jointly). Every
reported probability and mean is accompanied by its replication count $R$.

\paragraph{Verification.}
Before any result is produced, the implementation passes a ten-point test
contract (row-stochasticity of $W$ and $B$; bounded confidence; bounded values;
the simultaneous-update identity; seed reproducibility; the no-social baseline;
isolated communities $B=I$; full mixing; exactness of the quotient field; and
the self-weighting convention), together with the drift--diffusion validation
and an exact agreement check between the fast block engine and a dense-$W$
reference engine. Appendix~\ref{app:tests} lists the tests.

\paragraph{Reproducibility and scale.}
All randomness is seeded. Results below are reported at a \emph{standard}
compute scale ($N=400$ agents per two-community society, $120$--$300$
replications per configuration, $13\times13$ phase grids); the replication
package also ships a \emph{publication} scale ($N=1000$, up to $1000$
replications, $21\times21$ grids) with identical mechanism parameters. The
default parameters are listed in Appendix~\ref{app:params}.

\FloatBarrier
\section{Results}
\label{sec:results}

\FloatBarrier
\subsection{Illustrative trajectories of the collective regimes}
\label{sec:results-baseline}
\begin{figure}[H]
\centering
\includegraphics[width=\textwidth]{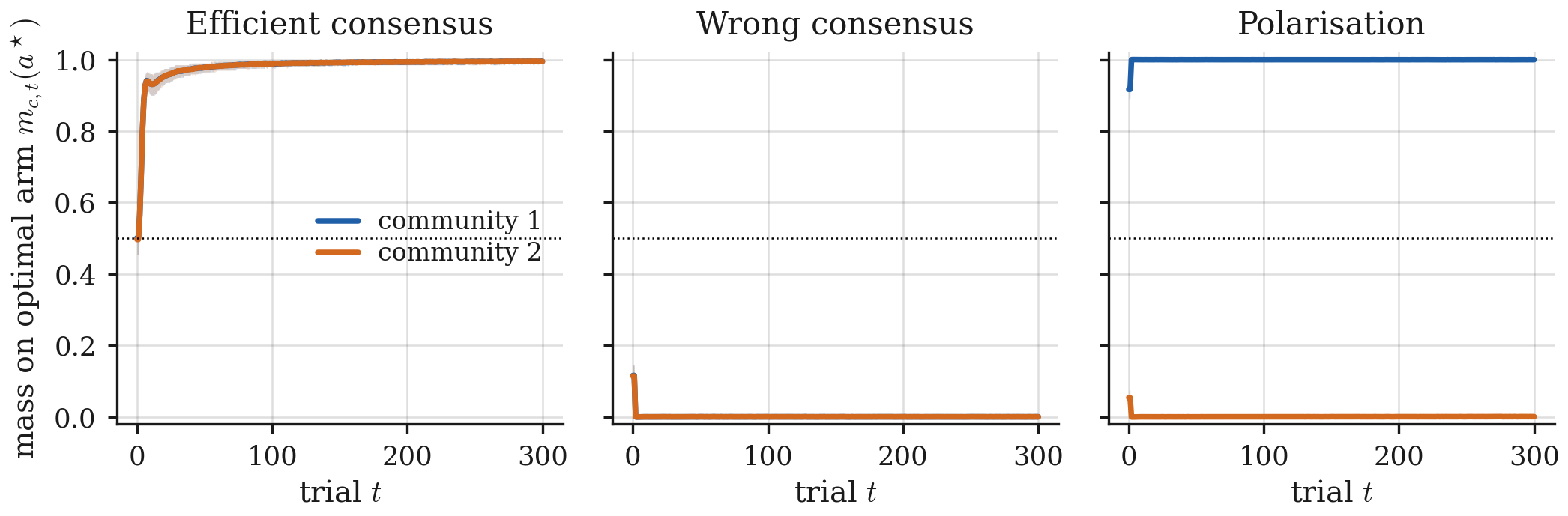}
\caption{Illustrative baseline regimes. Mass on the optimal arm $m_{c,t}(a^\star)$ in each
community (mean and $10$--$90$ percentile band over $R=300$ Monte Carlo
replications; $N=400$ agents, horizon $T=300$). The same mechanism produces
efficient consensus, wrong consensus and polarisation under the scenario
parameters of Table~\ref{tab:scenarios}.}
\label{fig:baseline}
\end{figure}

Figure~\ref{fig:baseline} provides three illustrative trajectories selected to show that the same behavioural architecture can support efficient consensus, wrong consensus and polarisation under different parameter configurations. These scenarios are descriptive examples rather than the basis of the comparative-static claims below. In the efficient configuration, both communities converge on the optimal arm ($\Prob=1.00$, mean regret $450$); in the wrong-consensus configuration, both lock on the suboptimal arm ($\Prob=1.00$, regret $\mathbf{11{,}988}$); and in the low-permeability configuration, the communities remain polarised ($\Prob=1.00$, regret $5{,}998$). Each illustrative regime occurs in all $R=300$ replications for its stated scenario (Wilson $95\%$ lower bound $\ge0.987$; Table~\ref{tab:baseline}). The economically relevant comparison is that shared error can generate substantially greater regret than persistent disagreement.

\begin{table}[H]
\centering
\small
\caption{Illustrative baseline regimes. Each scenario is run for $R$ Monte Carlo
replications; the empirical frequency of the designated regime is $1.00$ in every
case, reported with its Wilson $95\%$ lower bound. $\mathcal R_T$ is mean group
regret with Monte Carlo standard error. Scenario parameters are in
Table~\ref{tab:scenarios}.}
\label{tab:baseline}
\begin{tabular}{lcccc}
\toprule
Regime & $R$ & $\lambda$ & $P(\mathrm{regime})$ [Wilson $95\%$ LB] & $\mathcal R_T$ (SE) \\
\midrule
Efficient  & 300 & 0.40 & 1.00 [0.987] & 450 (1) \\
Wrong      & 300 & 0.90 & 1.00 [0.987] & 11,988 (0) \\
Polarised  & 300 & 0.90 & 1.00 [0.987] & 5,998 (0) \\
\bottomrule
\end{tabular}
\end{table}
\FloatBarrier

\subsection{Social transmission is non-monotone: the phase diagram}
\label{sec:results-phase}
\begin{figure}[H]
\centering
\includegraphics[width=\textwidth]{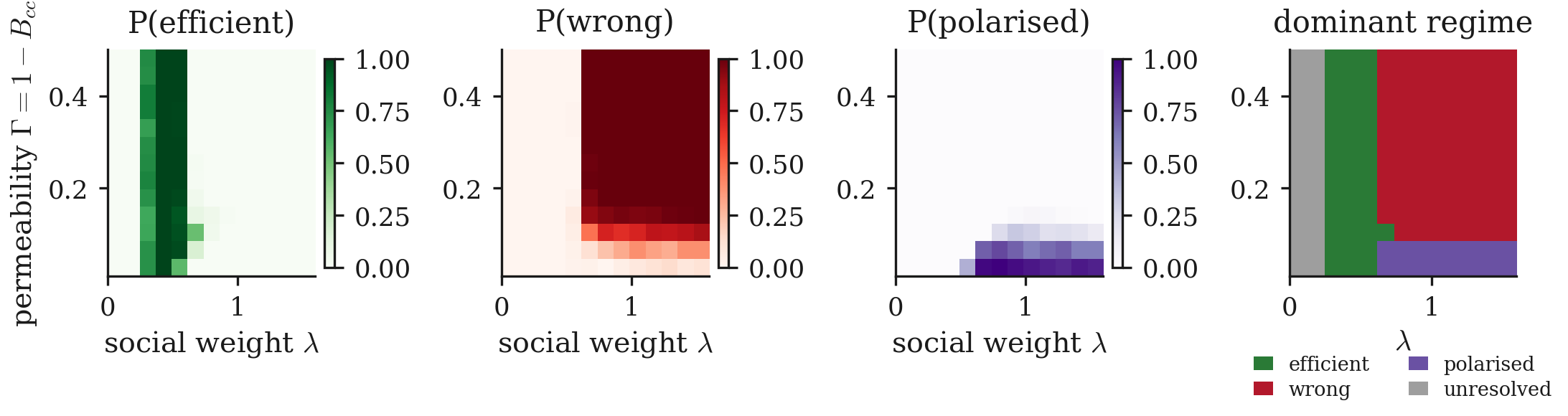}
\caption{Social-transmission phase diagram. Probability of efficient consensus, wrong consensus and polarisation, and the dominant regime, over the anticipatory social-transmission intensity \(\lambda\) and cross-community permeability \(\Gamma=1-B_{cc}\). A corrective band at moderate \(\lambda\) separates an unresolved region (\(\lambda\to0\)) from a distortive region (large \(\lambda\)); within the distortive region, low permeability yields polarisation and high permeability yields wrong consensus. Grid \(13\times13\), 120 replications per cell, regime threshold \(0.9\), horizon \(T=240\); the first three panels show regime probabilities and the fourth the dominant (modal) regime. The Wilson half-width at 120 replications is at most \(0.09\).}
\label{fig:phase}
\end{figure}

The central result is the phase diagram in Figure~\ref{fig:phase}, which varies two conceptually distinct objects: $\lambda$, the intensity of anticipatory social transmission, and $\Gamma$, cross-community permeability. The figure therefore separates the strength of social influence from the degree to which influence crosses community boundaries. Three findings stand out. First, collective performance is non-monotone in $\lambda$. At $\lambda\approx0$, learning is too slow for many populations to reach the consensus criterion within the horizon. At moderate $\lambda\in[0.27,0.53]$, social transmission accelerates correction and efficient consensus dominates across the permeability range. At larger $\lambda$, early confidence-weighted differences are increasingly amplified and inefficient outcomes emerge. Second, permeability determines the form of that amplification. High permeability allows an initially wrong community to transmit its lead across the system, making wrong consensus more likely; low permeability insulates communities and makes persistent polarisation more likely. This separation is consistent with Corollary~\ref{cor:twomodes}: permeability leaves the common-mode multiplier unchanged but attenuates the anti-symmetric mode by $1-2\Gamma$, so higher permeability locally suppresses relative community differences compared with common movement. Third, across the reported grid, wrong consensus is the modal outcome in $51\%$ of the $(\lambda,\Gamma)$ cells, efficient consensus in $24\%$, polarisation in $9\%$ and unresolved outcomes in $15\%$. The informational value of social transmission is therefore conditional on its intensity, while network permeability governs how amplified errors propagate across communities.

\FloatBarrier
\subsection{A dynamical-systems view: bifurcation and phase portrait}
\label{sec:dynamics}
To interpret the local threshold dynamically, we study a deterministic representative-community reduction on balanced blocks (\texttt{credibility/meanfield.py}). Within a community the reduction replaces individual stochastic choices by the DDM choice probability $\Lambda(2\beta a\Delta/\sigma^2)$, decision times by their mean first-passage value and Bernoulli rewards by their expectations. This is an analytical approximation to the aggregate stochastic dynamics, not an exact representation of the full value process. Freezing values at the indifference point isolates the object of Proposition~\ref{prop:amplification}: anticipatory feedback on the population lead $z_c=m_c(0)-\tfrac12$. The common mode has gain $\rho_+=C_0(\beta a/\sigma^2)\lambda$ and crosses one at $\lambda^\star=\sigma^2/(\beta a C_0)$. In the symmetric two-community benchmark, the anti-symmetric mode has gain $\rho_-=\rho_+(1-2\Gamma)$, so permeability changes the relative persistence of community-specific leads without moving the common-mode threshold.

\begin{figure}[H]
\centering
\includegraphics[width=0.72\textwidth]{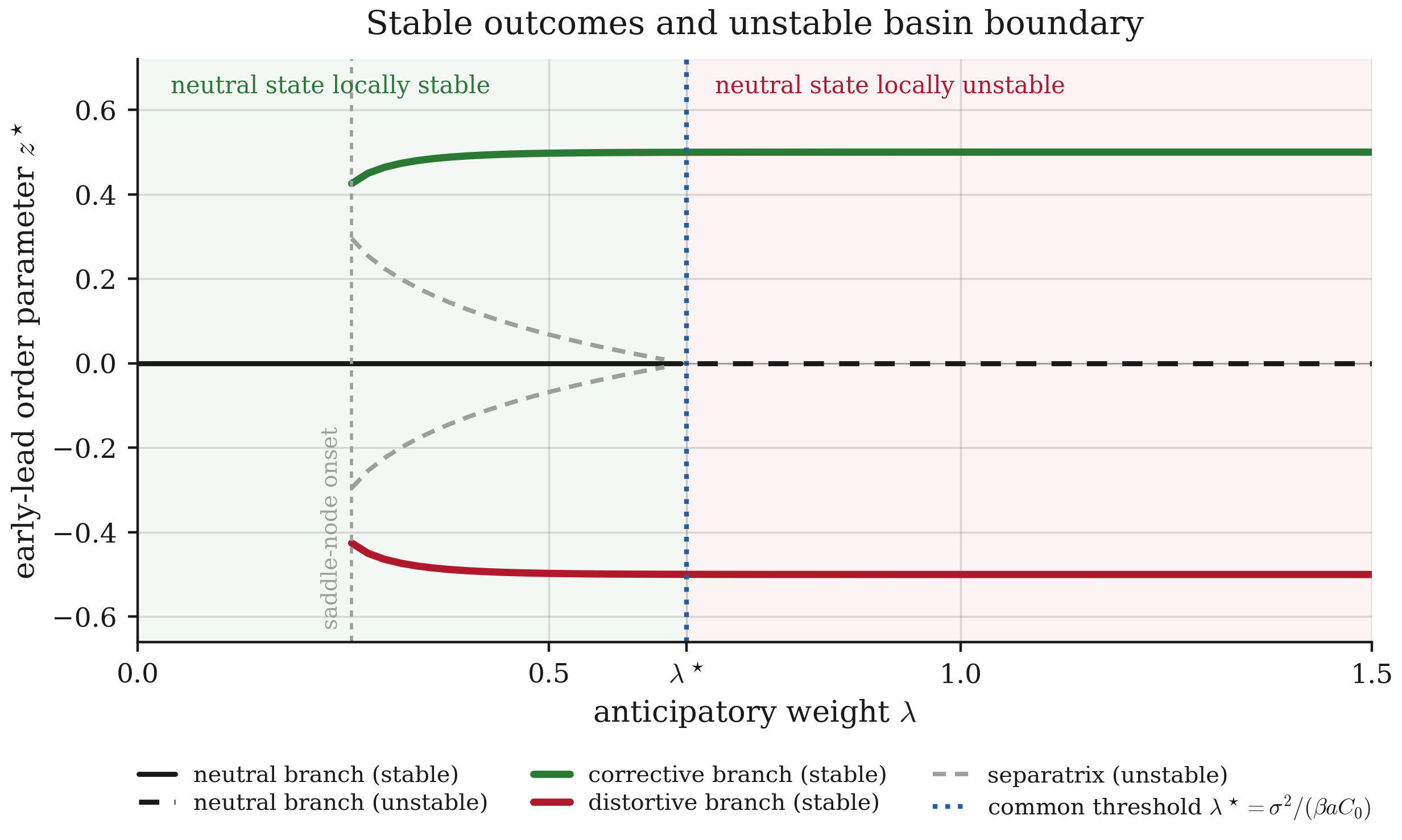}
\caption{Amplification bifurcation of the representative-community reduction. The early-lead
order parameter $z^\star$ is zero (stable neutral state) below $\lambda^\star$
and splits into a correct ($+z^\star$) and a wrong ($-z^\star$) amplified branch
above it. The transition is mildly subcritical (a bistable window precedes
$\lambda^\star$), so a sufficiently large early lead can lock in just below
threshold. Small reward gap $|\Delta\mu|=0.06$, $B_{cc}=0.85$.}
\label{fig:bifurcation}
\end{figure}

\begin{figure}[H]
\centering
\includegraphics[width=0.96\textwidth]{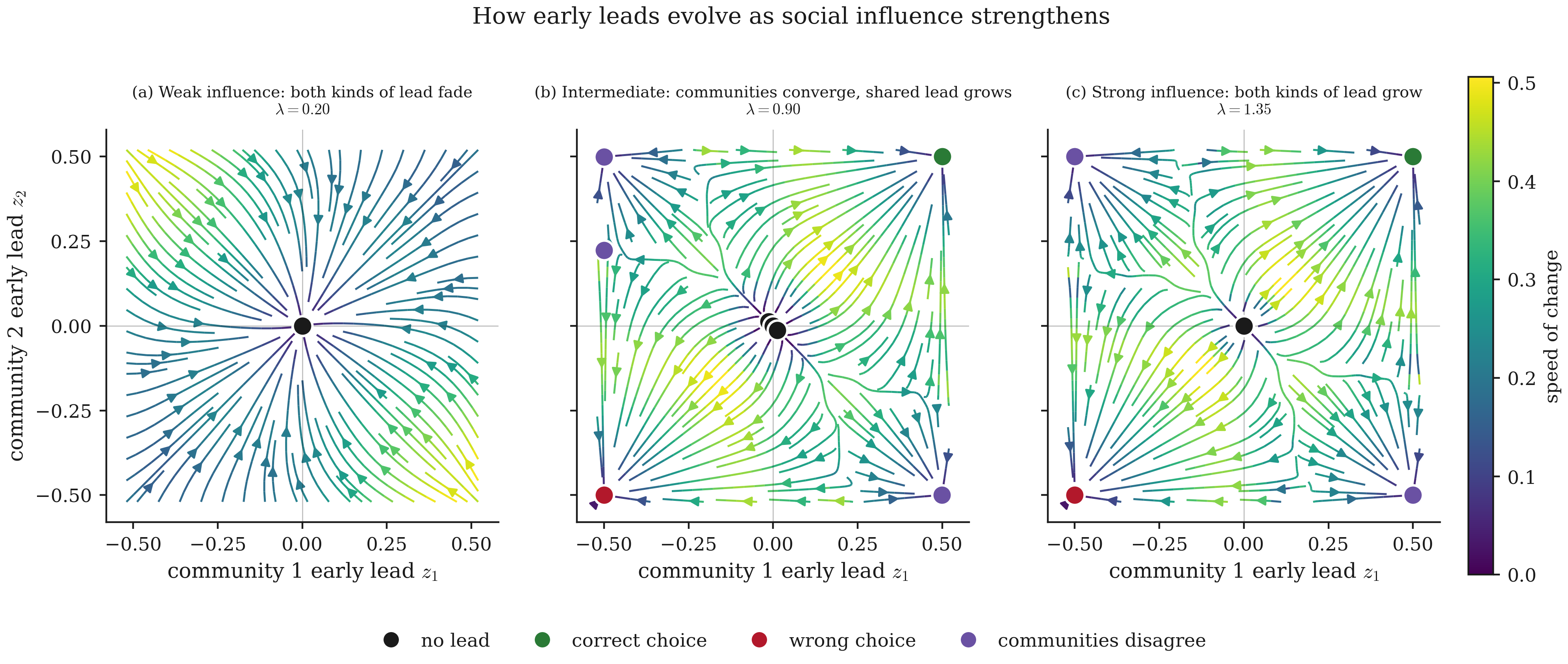}
\caption{How early community leads evolve as social influence strengthens.
Streamlines show the direction of motion, colour indicates its speed, and dots
mark fixed points. \textbf{(a)} Under weak influence, shared and opposing leads
both decay toward neutrality. \textbf{(b)} At intermediate influence, shared
leads grow while opposing leads decay, producing consensus. \textbf{(c)} Under
strong influence, both modes grow, permitting corrective $(+,+)$, distortive
$(-,-)$ and polarised $(\pm,\mp)$ endpoints. The three panels therefore isolate
the modal information not contained in the scalar bifurcation diagram.
Frozen-value reduction, $|\Delta\mu|=0.06$, $B_{cc}=0.85$.}
\label{fig:flowfield}
\end{figure}

Within this reduced system, Figure~\ref{fig:bifurcation} illustrates the loss of local stability at the amplification threshold. Below $\lambda^\star$, the neutral state is locally stable and small leads decay; above it, the neutral state loses stability and amplified branches emerge. Figures~\ref{fig:flowfield} provide complementary phase-space representations of the same reduced dynamics. They show how the basin structure changes once anticipatory feedback becomes locally amplifying and how the anti-symmetric community mode can support polarised outcomes. These diagrams should be read as an analytical account of the feedback mechanism, not as a proof of the terminal regime of the finite-$N$ stochastic model. The latter is measured directly by the Monte Carlo experiments in Figure~\ref{fig:phase}, and the approximation is evaluated against the agent model in Section~\ref{sec:results-quotient}.

\FloatBarrier
\subsection{Mechanism decomposition: amplification and buffering}
\label{sec:results-ablation}
\begin{figure}[H]
\centering
\includegraphics[width=\textwidth]{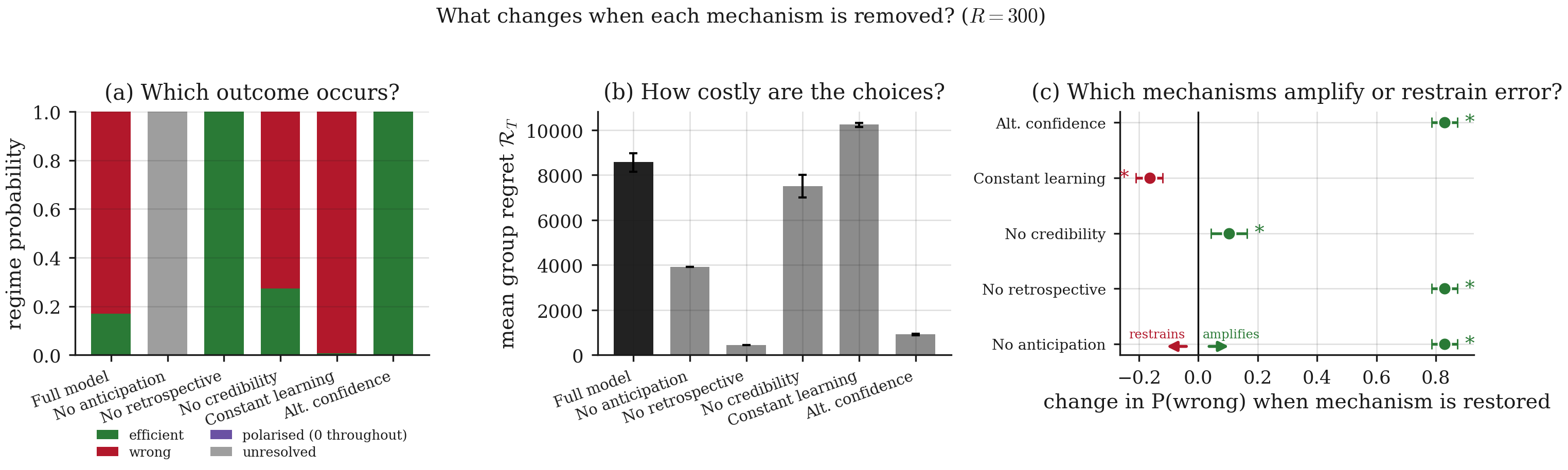}
\caption{Mechanism isolation under common random numbers (\(R=300\)). (a) Regime composition for the full model and five ablations at the contested operating point (Table 5); (b) mean group regret with \(95\%\) bootstrap intervals; (c) paired structural contrast in wrong-consensus incidence, \(\Delta P(\mathrm{wrong})=P(\mathrm{wrong}\mid\mathrm{full})-P(\mathrm{wrong}\mid\mathrm{ablation})\), with paired-bootstrap \(95\%\) intervals. Positive contrasts indicate that removing the component lowers wrong-consensus incidence; negative contrasts indicate that removing it raises wrong-consensus incidence. At this operating point, removing the anticipatory channel, the retrospective channel, or the decision-time confidence map eliminates wrong consensus; removing credibility weighting weakens but does not eliminate it; constant learning rates intensify it.}
\label{fig:ablations}
\end{figure}

We decompose the mechanism using the paired structural counterfactuals of Section~\ref{sec:identification}. The full model and five ablations are run under common random numbers ($R=300$) at a contested operating point where the full model produces wrong consensus with probability $0.83$ (Wilson $95\%$ interval $[0.78,0.87]$). Figure~\ref{fig:ablations} and Table~\ref{tab:ablation} report regime frequencies and paired component contrasts. The results separate three roles. First, at this operating point, removing the anticipatory channel ($\lambda=0$), the retrospective channel ($\eta=0$), or the decision-time dependence of confidence drives wrong consensus to $0.00$; these elements are therefore necessary for the observed distortion at this configuration. Second, removing confidence-sensitive credibility weighting weakens but does not eliminate wrong consensus ($0.83\!\to\!0.73$, paired contrast $+0.10$, $[0.04,0.16]$). Credibility weighting is thus an amplifier rather than the sole source of the failure. Third, holding private learning rates constant intensifies wrong consensus ($0.83\!\to\!0.99$, paired contrast $-0.16$, $[-0.21,-0.12]$). Confidence therefore has a dual behavioural role: it strengthens the social propagation of selected signals while the confidence-modulated private learning rule provides a countervailing correction mechanism after confident mistakes. The inefficient regime is best understood as the outcome of this feedback architecture, not as the consequence of any single parameter in isolation.

\begin{table}[H]
\centering
\small
\caption{Mechanism isolation at the contested operating point
(Table~\ref{tab:scenarios}), $R=300$ replications under common random numbers.
$P(\mathrm{wrong})$ and $P(\mathrm{efficient})$ carry Wilson $95\%$ intervals;
regret carries a $95\%$ bootstrap interval. The last column reports the paired
structural contrast
$\Delta=P(\mathrm{wrong}\mid\mathrm{full})-P(\mathrm{wrong}\mid\mathrm{ablation})$,
with a paired-bootstrap $95\%$ interval. A positive $\Delta$ indicates that
removing the component reduces wrong-consensus incidence, whereas a negative
$\Delta$ indicates that removing it increases wrong-consensus incidence. A
component is described as necessary at this operating point only when its
removal reduces $P(\mathrm{wrong})$ to approximately zero.}
\label{tab:ablation}
\resizebox{\textwidth}{!}{%
\begin{tabular}{lcccc}
\toprule
Variant & $P(\mathrm{wrong})$ [$95\%$] & $P(\mathrm{efficient})$ [$95\%$] & $\mathcal R_T$ [$95\%$] & $\Delta P(\mathrm{wrong})$ [$95\%$] \\
\midrule
Full & 0.83 [0.78, 0.87] & 0.17 [0.13, 0.22] & 8,571 [8,148, 8,962] & --- \\
No anticipatory & 0.00 [0.00, 0.01] & 0.00 [0.00, 0.01] & 3,913 [3,908, 3,919] & +0.83 [+0.79, +0.87] \\
No retrospective & 0.00 [0.00, 0.01] & 1.00 [0.99, 1.00] & 438 [433, 442] & +0.83 [+0.79, +0.87] \\
No credibility weighting & 0.73 [0.67, 0.77] & 0.27 [0.23, 0.33] & 7,501 [7,002, 7,996] & +0.10 [+0.04, +0.16] \\
Constant learning rates & 0.99 [0.98, 1.00] & 0.01 [0.00, 0.02] & 10,239 [10,142, 10,303] & $-0.16$ [$-0.21$, $-0.12$] \\
Alternative confidence & 0.00 [0.00, 0.01] & 1.00 [0.99, 1.00] & 911 [882, 941] & +0.83 [+0.79, +0.87] \\
\bottomrule
\end{tabular}%
}
\end{table}
\FloatBarrier

\subsection{Confidence attaches to error}
\label{sec:results-confidence}
\begin{figure}[H]
\centering
\includegraphics[width=\textwidth]{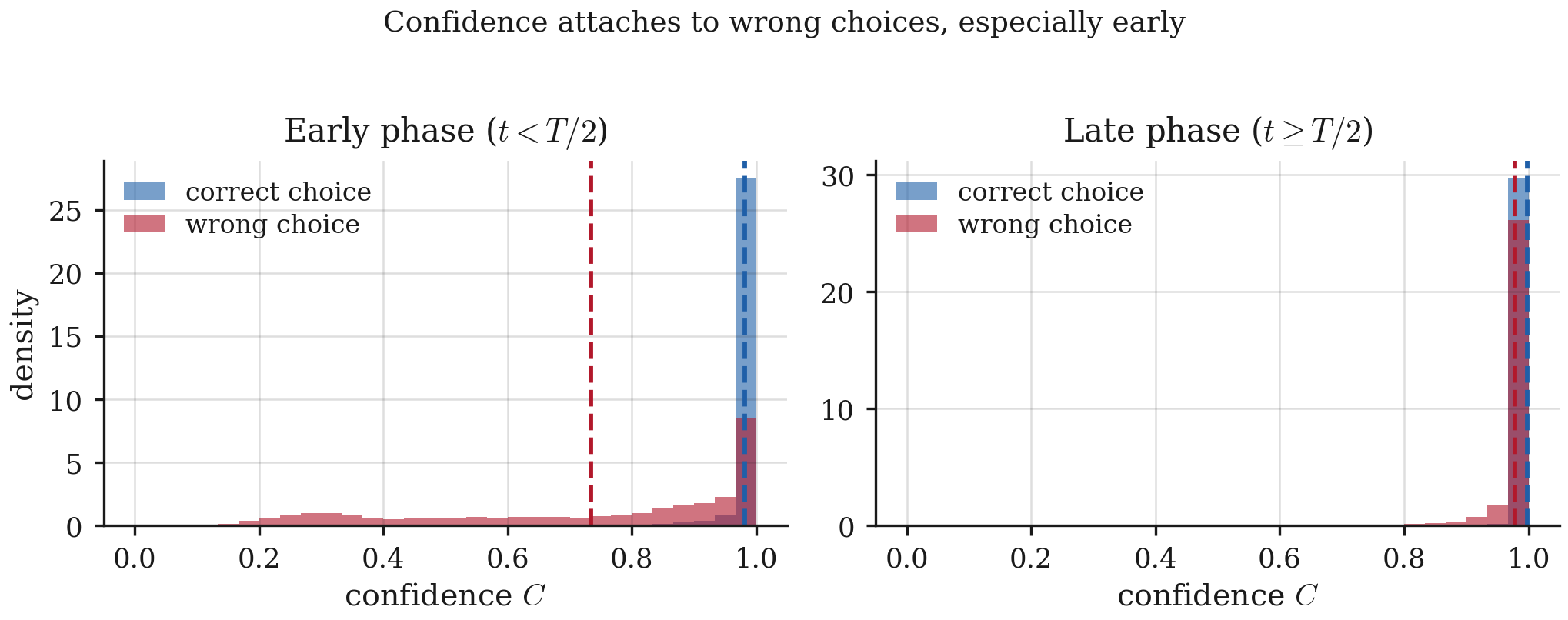}
\caption{Decision-process confidence and early confident error. Distribution of
confidence $C$ for correct and wrong choices, early ($t<T/2$) and late
($t\ge T/2$), in a neutral contested scenario; densities are normalised to unit
area and dashed lines mark group means. Early decisions number $\approx1.8\times10^6$
(correct) and $\approx6\times10^4$ (wrong). Early wrong choices carry a wide
range of confidence, including high values (mean $0.73$ versus $0.98$ for correct
choices): confidence attaches to error, which is what lets credibility-weighted
transmission amplify it. The right panel shows that confidence concentrates near
one late in learning as communities reach consensus.}
\label{fig:confidence}
\end{figure}

Why can transmission distort? Because confidence is generated by the decision process and can attach to
wrong choices early in learning, exactly as the mechanism requires.
Figure~\ref{fig:confidence} shows the distribution of confidence conditional on
whether the chosen arm was optimal, split by phase. Early in learning, wrong choices carry substantial confidence
(mean $0.73$, with appreciable mass above $0.9$), only modestly below the
confidence of correct choices (mean $0.98$). These early confident errors are
precisely the signals that the anticipatory channel amplifies before enough
corrective evidence has accumulated. The mechanism requires no global
irrationality --- only that early stochastic success can generate confidence,
and that confidence is socially read as credibility.

\FloatBarrier
\subsection{The community quotient is faithful in the bulk and breaks at boundaries}
\label{sec:results-quotient}
\begin{figure}[H]
\centering
\includegraphics[width=\textwidth]{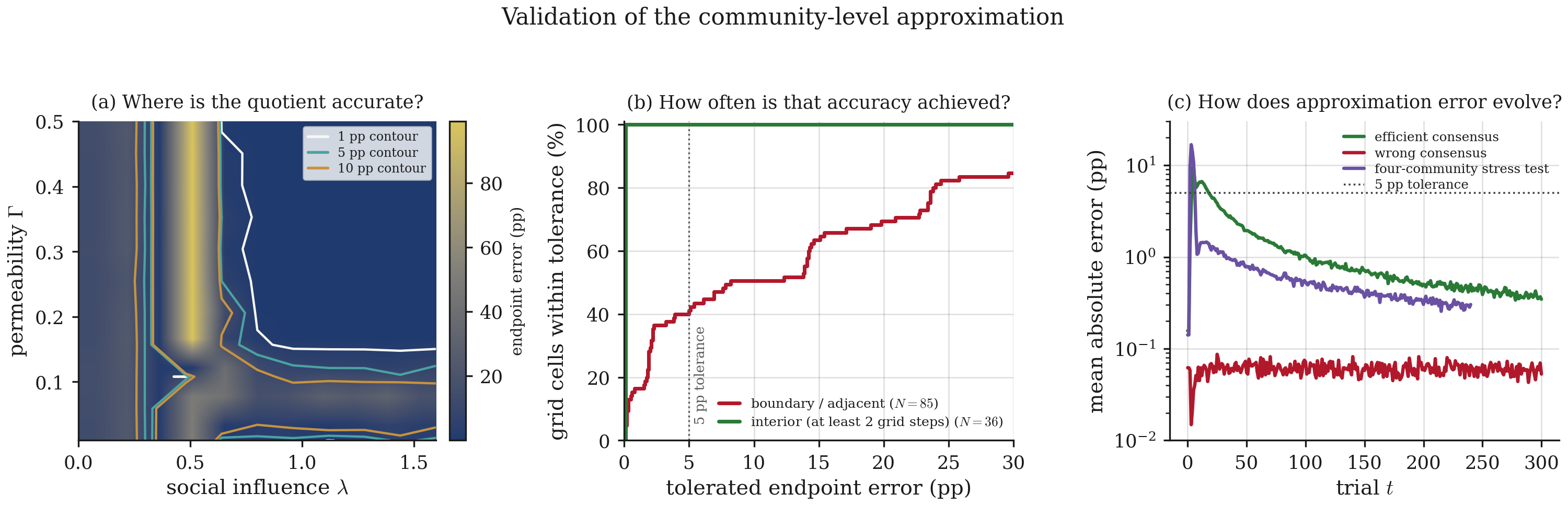}
\caption{Validity of the community-level approximation. \textbf{(a)} Mean
absolute endpoint error over the $11\times11$ $(\lambda,\Gamma)$ grid, with
contours delimiting one-, five- and ten-percentage-point tolerances.
\textbf{(b)} Empirical share of parameter cells meeting each tolerance,
stratified by distance from an agent-model regime boundary. All 36 interior
cells are within one percentage point; errors are concentrated among the 85
boundary or adjacent cells. \textbf{(c)} Mean absolute micro--meso error over
time for efficient consensus, wrong consensus and a four-community stress test;
the logarithmic scale distinguishes small persistent discrepancies from large
but transient ones. The dotted guides in panels (b) and (c) mark a
five-percentage-point tolerance.}
\label{fig:quotient}
\end{figure}

Finally we validate the community-level (quotient) description against the full
agent model (Figure~\ref{fig:quotient}). In two- and four-community examples the
deterministic meso recursion tracks the micro ensemble mean closely, including a
modular four-community society in which both descriptions reach the optimal arm
(terminal micro mass $0.997$ per community versus meso $1.000$). Across a
$(\lambda,\Gamma)$ sweep the mean absolute discrepancy between the micro
ensemble and the meso trajectory is $0.137$, and the two descriptions agree on
the dominant regime in $79\%$ of cells. Consistent with
Remark~\ref{rem:exact}, the discrepancy is small in the interior of regimes and
concentrates along the corrective/distortive phase boundary near
$\lambda\approx0.5$, where finite-$N$ fluctuations decide the regime and the
deterministic reduction is least reliable. The quotient is thus a useful but
bounded instrument: exact for the anticipatory field, accurate for aggregate
behaviour away from transitions, and explicitly not a substitute for the
agent-level model near them.

\FloatBarrier
\subsection{Robustness}
\label{sec:results-robustness}
The phase structure is robust over the sensitivity checks reported in Appendix~\ref{app:robust}, Table~\ref{tab:robustness} and Figure~\ref{fig:robustness}. The efficient, wrong-consensus and polarised regions persist across terminal thresholds ($0.80/0.90/0.95$), horizons up to $T=1000$, reward gaps $|\Delta\mu|\in[0.02,0.20]$, and moderate departures from balanced block weights using weighted stochastic-block and noisy row-normalised graphs. The dominant-regime map is also stable across the reported phase-grid resolutions and under the deterministic reaction-time approximation. In particular, the unresolved region near $\lambda=0$ is not an artefact of the baseline horizon: it remains visible at $T=1000$. The principal sensitivity is to the size of the early lead, as expected from Proposition~\ref{prop:amplification}; at the distortive operating point an initial asymmetry of $\delta=0.04$ is already sufficient to produce wrong consensus. Residual uncertainty is concentrated near the transition region, where finite-$N$ fluctuations and the deterministic community approximation are least reliable.


\FloatBarrier
\section{Discussion}
\label{sec:discussion}

The results support four main conclusions.

\emph{The informational value of social transmission is non-monotone.} The relevant comparative static is the anticipatory influence intensity $\lambda$, not network connectivity itself. Weak transmission leaves learning slow over the experimental horizon, moderate transmission accelerates correction, and sufficiently strong transmission makes early confidence-weighted differences self-reinforcing. The analytical threshold provides a local explanation for this transition, while the phase diagram establishes its global expression in the stochastic model.

\emph{Confidence has both amplifying and stabilising roles.} Confidence is generated by the decision process rather than imposed as an exogenous signal. When used socially, it increases the effective exposure to selected neighbours; when used privately, it changes the gain applied to prediction errors. The ablations show that these roles need not point in the same direction: credibility-sensitive transmission increases wrong-consensus risk at the contested point, whereas confidence-dependent private learning reduces it. Process-generated confidence is therefore not simply an ``amplifier''; it is a state variable that reallocates both social influence and individual correction.

\emph{Network permeability selects the form of collective failure.} Low cross-community permeability limits the spread of a confident error from one community to another, but the same insulation can preserve disagreement. High permeability instead makes community-level errors more likely to become shared. Corollary~\ref{cor:twomodes} provides the local analytical counterpart: for two symmetric communities the common mode is unaffected by $\Gamma$, while the anti-symmetric mode is scaled by $1-2\Gamma$. Modularity therefore trades off common mislearning against persistent disagreement rather than uniformly improving collective accuracy; the nonlinear stochastic simulations determine which basin is ultimately reached.

\emph{Mesoscopic aggregation is useful but deliberately bounded.} Proposition~\ref{prop:quotient} gives an exact representation of the anticipatory social field under balanced block weights, but the deterministic representative-community dynamics remain an approximation to the stochastic value process. Their role is analytical: they expose the feedback threshold and organise the phase portrait. The agent-level simulations remain the relevant object for regime probabilities, especially close to transition boundaries.

\paragraph{Cognitive interpretation.}
The parameters carry direct cognitive readings: $\beta$ is value sensitivity
(how sharply value differences drive choice), $\sigma$ internal decision noise,
$a$ the speed--accuracy threshold, $\kappa_1,\kappa_2$ the evidence- and
time-dependence of confidence, $\alpha$ the reward learning rate, $\lambda$
susceptibility to social influence, and $\eta$ retrospective credit assignment
(Appendix~\ref{app:params}). The retrospective channel is an other-referenced
prediction error, the social counterpart of the reward prediction error
\citep{JoinerPivaTurrinChang2017}, and the confidence-modulated learning rule is
a teaching-like signal in the sense of orbitofrontal adaptive-learning accounts
\citep{SchuckCaiWilsonNiv2016,Niv2019,WangVeismannBanerjeePleger2023}. The model
tracks action values rather than latent task states; relating credibility
transmission to belief-state or cognitive-map inference
\citep{Niv2019,BehrensMullerWhittingtonMarkSchultzeDolanKurthNelson2018} is a
natural theoretical extension.

\paragraph{Welfare relative to private learning.}
Group regret provides a direct economic benchmark for the regime classification. Relative to the same agents with $\lambda=\eta=0$, social learning has a sign-changing welfare effect. At the corrective operating point, mean regret falls from about $7{,}400$ under private learning to $450$ under the social model. At the distortive point, mean regret rises from about $4{,}800$ to roughly $12{,}000$. The change from corrective to distortive social learning is therefore economically meaningful: across these matched benchmark comparisons, the welfare value of social information reverses sign relative to private reinforcement learning.

\paragraph{Testable predictions.}
The mechanism yields four direct experimental predictions.\\ \\
\textbf{(i) Confidence conditional on accuracy:} a receiver's response to a sender should increase with the sender's decision confidence even after controlling for whether the sender is correct. \\
\textbf{(ii) Persistence of early confident error:} conditional on the same objective information, an early mistake expressed with high confidence should have a larger and more persistent effect on subsequent group behaviour than the same mistake expressed with low confidence. \\
\textbf{(iii) Decision uncertainty and response time:} raising uncertainty, or otherwise lowering confidence, should weaken social transmission; since Eq.~\ref{eq:confidence} makes confidence decreasing in decision time at matched evidence, the model predicts less following of slower, otherwise comparable senders. \\
\textbf{(iv) Exogenous confidence manipulation:} experimentally shifting confidence while holding signal accuracy as fixed as possible---for example through speed pressure, evidence-quality framing, or an explicit confidence display---should change the strength of social transmission. These predictions map the model onto observables that can distinguish credibility-sensitive social learning from an account in which receivers react only to sender accuracy.

\begin{figure}[H]
\centering
\includegraphics[width=\textwidth]{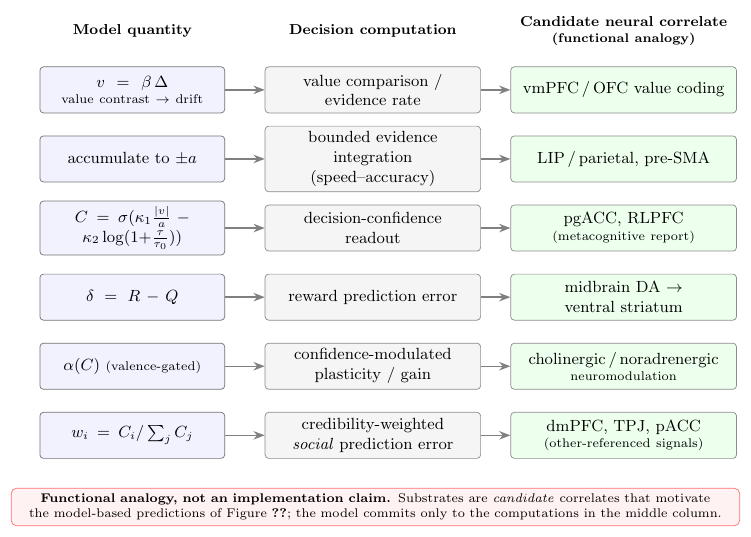}
\caption{Computational-to-neural correspondence. Each model quantity (left)
implements a decision computation (centre) with a candidate neural correlate
(right). A functional analogy: it motivates the measurements and manipulations of
Figure~\ref{fig:experiment}; the model commits to the centre column.}
\label{fig:neuromap}
\end{figure}

\begin{figure}[H]
\centering
\includegraphics[width=\textwidth]{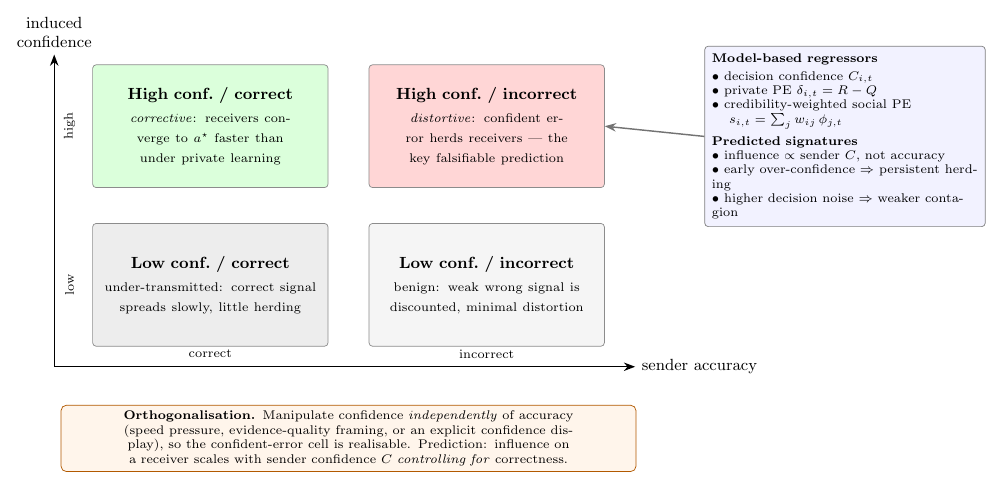}
\caption{A model-based experimental design. Inducing confidence orthogonally to
accuracy realises the confident-error cell (upper right), where the model
predicts distortive transmission. The model supplies the regressors and
falsifiable signatures (right), turning predictions \emph{(i)--(iv)} into a
testable protocol for dyadic or networked group-decision experiments.}
\label{fig:experiment}
\end{figure}

\paragraph{A neuroeconomic reading.}
The model is compatible with a process-based interpretation in which value comparison drives evidence accumulation, decision time contributes to confidence, and reward prediction errors update values \citep{JoinerPivaTurrinChang2017}. Figure~\ref{fig:neuromap} summarises candidate computational-to-neural correspondences, but these mappings are auxiliary: the paper commits to the behavioural computations, not to particular neural substrates. Their main use here is to motivate the measurable quantities in the experimental design of Figure~\ref{fig:experiment}. In particular, the proposed design orthogonalises confidence and accuracy and tests whether receiver updating scales with sender confidence conditional on correctness.

\paragraph{Limitations and scope.}
The baseline is deliberately narrow: binary actions, stationary rewards, fixed networks, community-homogeneous parameters and a progression from two communities to a modular ring. Confidence is treated as socially available, whereas in many settings it must be inferred from reaction time, hesitation, explicit reports or past reliability. The anticipatory and retrospective channels are also modelled separately, and network weights are fixed rather than reputation-dependent. These restrictions are useful because they isolate the decision-generated-credibility feedback, but they delimit the claims: the model is a synthetic mechanism model, not a calibrated description of a specific empirical population. The experimental predictions above provide the natural route for testing which of these restrictions should be relaxed in future work.

\FloatBarrier
\section{Conclusion}
We study social learning under \emph{decision-generated credibility}: the weight attached to a social signal is produced by the same decision process that generates the transmitted action. In the model, reinforcement learners choose through a drift--diffusion process, confidence is read out from that process, and confidence subsequently changes both social exposure and learning gains. The resulting feedback has a simple local threshold: below it, small confidence-weighted social differences are damped; above it, they are amplified.

The full stochastic model shows why this matters economically. Social-transmission intensity has a non-monotone relationship with collective performance: moderate transmission accelerates correction, whereas strong transmission can transform early confident errors into persistent wrong consensus. Cross-community permeability determines whether amplification spreads across communities or remains local and appears as polarisation. Paired component ablations further show that confidence plays opposing roles across channels---it can strengthen social propagation while also increasing private correction after confident mistakes.

The central implication is therefore not that social connection is intrinsically beneficial or harmful. Rather, the informational value of social influence depends on the behavioural process that determines how strongly social signals are weighted. This mechanism yields direct empirical predictions: receivers should respond to sender confidence conditional on sender accuracy, and early high-confidence errors should exert disproportionately persistent social effects.

\FloatBarrier
\section*{Code and data availability}
All code, configurations, fixed seeds and analysis scripts that reproduce every
figure and table are released as the open-source \textsc{csl} (Cognitive Social
Learning) repository under an MIT license:
\url{https://github.com/TrompetteMarine/CSL}. The repository documents the verification suite, the
common-random-number ablation protocol, and one-command reproduction of the
standard and publication profiles.

\appendix

\FloatBarrier
\section{Analytical results and proofs}
\label{app:proofs}

This appendix supplies the derivations behind the analytical statements in the
main text. The two probabilistic inputs used for the drift--diffusion identities
--- optional stopping for the scale function and Dynkin's equation for the mean
exit time --- are standard results from stochastic-process theory
\citep{KaratzasShreve1991}. All subsequent algebraic, analytic and finite-network
steps are derived below.

\FloatBarrier
\subsection{Drift--diffusion first-passage identities}
\label{app:ddmproof}

Let $X_s=vs+\sigma B_s$ start at $0$ and let
$\tau=\inf\{s\ge0:|X_s|=a\}$. For $v\neq0$, the scale function
$s(x)=\exp(-2vx/\sigma^2)$ makes $s(X_{s\wedge\tau})$ a bounded stopped
local martingale. Optional stopping therefore gives
\[
1=s(0)=\E[s(X_\tau)]
=p\,e^{-2va/\sigma^2}+(1-p)e^{2va/\sigma^2},
\]
where $p=\Prob(X_\tau=a)$. Since $v\neq0$ implies
$e^{-2va/\sigma^2}\neq e^{2va/\sigma^2}$, the linear equation has the unique
solution
\[
p=\frac{1}{1+e^{-2va/\sigma^2}}.
\]
At $v=0$, symmetry gives $p=1/2$, which is also the continuous limit.

For the mean exit time, $u(x)=\E_x[\tau]$ solves the boundary-value problem
\[
\frac{\sigma^2}{2}u''(x)+v u'(x)=-1,
\qquad u(-a)=u(a)=0,
\]
by Dynkin's equation. For $v\neq0$ the general solution is
$u(x)=-x/v+A+B e^{-2vx/\sigma^2}$. Imposing both boundary conditions and
evaluating at $x=0$ gives
\[
\bar\tau(v)=u(0)=\frac{a}{v}\tanh\!\left(\frac{av}{\sigma^2}\right).
\]
Writing $z=av/\sigma^2$,
\[
\bar\tau(v)=\frac{a^2}{\sigma^2}\frac{\tanh z}{z},
\]
so $\bar\tau(v)\to a^2/\sigma^2$ as $v\to0$ because
$\tanh z/z\to1$.

Several properties used in the confidence analysis follow immediately. First,
$\bar\tau$ is even because $\tanh$ is odd. Second, for $v\neq0$ it is positive
and
\[
0<\frac{\tanh |z|}{|z|}<1,
\qquad
0<\bar\tau(v)<\frac{a^2}{\sigma^2}.
\]
Third, $f(z)=\tanh z/z$ is strictly decreasing for $z>0$ because
\[
f'(z)=\frac{z\operatorname{sech}^2 z-\tanh z}{z^2}<0,
\]
and the numerator is negative exactly when
$\sinh z\cosh z>z$, equivalently $\sinh(2z)>2z$. Hence
$\bar\tau(v)$ is strictly decreasing in $|v|$.

\FloatBarrier
\subsection{Properties of process-generated confidence}
\label{app:confidenceproof}

For evidence magnitude $e\ge0$ and decision time $\tau\ge0$, write
\[
C(e,\tau)=\Lambda\!\left(\kappa_1\frac{e}{a}
-\kappa_2\log\!\left(1+\frac{\tau}{\tau_0}\right)\right).
\]
Since $0<\Lambda(x)<1$, confidence lies in $(0,1)$. Using
$\Lambda'(x)=\Lambda(x)[1-\Lambda(x)]$,
\[
\frac{\partial C}{\partial e}
=\frac{\kappa_1}{a}C(1-C)>0,
\qquad
\frac{\partial C}{\partial\tau}
=-\frac{\kappa_2}{\tau_0+\tau}C(1-C)<0.
\]
Thus the direct evidence and time channels have the stated signs.

Along the mean-decision-time ridge, Eq.~\eqref{eq:confridge} depends on
$\Delta$ only through $|\Delta|$ and the even function
$\bar\tau(\beta\Delta)$, so $\bar C(-\Delta)=\bar C(\Delta)$. For
$\Delta>0$, evidence magnitude $\beta\Delta$ rises while
$\bar\tau(\beta\Delta)$ strictly falls; both changes raise confidence. Hence
$\bar C$ is strictly increasing on $\Delta>0$ and therefore U-shaped in the
signed contrast.

At indifference, $|v|=0$ and
$\bar\tau(0)=a^2/\sigma^2$, giving
\[
C_0
=\Lambda\!\left[-\kappa_2\log\!\left(1+\frac{a^2}{\sigma^2\tau_0}\right)\right]
=\frac{1}{1+\left(1+a^2/(\sigma^2\tau_0)\right)^{\kappa_2}}.
\]
The logarithm is strictly positive, so its argument under $\Lambda$ is negative
and $0<C_0<1/2$.

Finally, $|\Delta|$ is Lipschitz at zero and $\bar\tau(\beta\Delta)$ is even
and continuous with a quadratic departure from its zero-drift value. Since the
logarithm and logistic map are continuously differentiable on the relevant
compact neighbourhood,
\[
\bar C(\Delta)=C_0+O(|\Delta|).
\]
The absolute-value evidence term generally gives non-zero and opposite
one-sided derivatives at zero, so the confidence map need not be differentiable
at indifference. This is why Proposition~\ref{prop:amplification} uses the
product expansion \eqref{eq:massgaplin} rather than differentiating confidence
itself.

The high-confidence-error symmetry is immediate. Let
$p(\Delta)=\Lambda(2\beta a\Delta/\sigma^2)$. Then
$p(-\Delta)=1-p(\Delta)$ but $\bar C(-\Delta)=\bar C(\Delta)$. Thus equal and
opposite subjective contrasts carry identical confidence while reversing the
choice tendency.

\FloatBarrier
\subsection{Learning-rate and state-space bounds}
\label{app:learningbounds}

Let
\[
\alpha^{-}(C)=\alpha^{\min}+(\alpha^{\max}-\alpha^{\min})C,
\qquad
\alpha^{+}(C)=\alpha^{\max}-(\alpha^{\max}-\alpha^{\min})C.
\]
For $C\in(0,1)$ and
$0<\alpha^{\min}\le\alpha^{\max}<1$, both rates lie in
$[\alpha^{\min},\alpha^{\max}]$, and
\[
\alpha^{-}(C)-\alpha^{+}(C)
=(\alpha^{\max}-\alpha^{\min})(2C-1).
\]
They therefore cross at $C=1/2$, with negative prediction errors receiving the
larger gain exactly when $C>1/2$.

For Eq.~\eqref{eq:socpe}, non-negativity of the weights and $0<C_j<1$ imply
\[
0\le
\sum_{j\neq i}W_{ij}\1\{A_j=a\}C_j
<
\sum_{j\neq i}W_{ij}\1\{A_j=a\}+\varepsilon,
\]
unless both sums vanish, in which case the ratio is zero. Hence
$0\le\bar C_{i,t}(a)<1$. With $\gamma>0$ and $\omega>0$,
$0\le g(\bar C)<\gamma$ and $g(0)=0$. Finally,
$\Pi_{[0,1]}$ maps every real pre-update value into $[0,1]$, so if the values
are initialised in $[0,1]$ then the value state space remains in $[0,1]^{2N}$
for all trials.

\FloatBarrier
\subsection{Balanced blocks, quotient aggregation and self-exclusion}
\label{app:quotientproof}

Under Assumption~\ref{ass:balanced}, every $N_d$ is strictly positive. For
$i\in V_c$,
\[
\sum_{j=1}^N W_{ij}
=\sum_{d=1}^M\sum_{j\in V_d}\frac{B_{cd}}{N_d}
=\sum_{d=1}^M B_{cd}=1,
\]
so row-stochasticity of $W$ is inherited from $B$. Similarly,
\[
b_{i\to d}=\sum_{j\in V_d}W_{ij}
=N_d\frac{B_{cd}}{N_d}=B_{cd}.
\]
For any statistic $f_j$,
\[
\sum_{j=1}^N W_{ij}f_j
=\sum_{d=1}^M\sum_{j\in V_d}\frac{B_{cd}}{N_d}f_j
=\sum_{d=1}^M B_{cd}\bar f_d,
\]
which proves Proposition~\ref{prop:quotient}; substituting
$f_j=\1\{A_{j,t-1}=a\}C_{j,t-1}$ gives
Eq.~\eqref{eq:quotientfield}.

If self is excluded,
\[
\sum_{j\neq i}W_{ij}f_j
=\sum_jW_{ij}f_j-W_{ii}f_i
=\sum_dB_{cd}\bar f_d-\frac{B_{cc}}{N_c}f_i.
\]
For $|f_i|\le1$ the absolute correction is therefore bounded by
$B_{cc}/N_c$, proving Corollary~\ref{cor:selfexclude}. Two useful limiting
cases are exact consequences: if $B=I$, communities decouple; if every row of
$B$ is uniform, all communities face the same quotient field.

\FloatBarrier
\subsection{Proof of the local network-mode amplification result}
\label{app:ampproof}

The logistic identity $2\Lambda(2x)-1=\tanh x$ gives
\[
h(\Delta)=2p(\Delta)-1=\tanh(\kappa\Delta),
\qquad \kappa=\frac{\beta a}{\sigma^2}.
\]
Since $\tanh x=x+O(x^3)$,
$h(\Delta)=\kappa\Delta+O(\Delta^3)$. From
Appendix~\ref{app:confidenceproof},
$\bar C(\Delta)=C_0+O(|\Delta|)$. Therefore
\[
\bar C(\Delta)h(\Delta)
=\bigl(C_0+O(|\Delta|)\bigr)
 \bigl(\kappa\Delta+O(\Delta^3)\bigr)
=C_0\kappa\Delta+O(\Delta^2),
\]
which proves Eq.~\eqref{eq:massgaplin} despite the kink in confidence at zero.
Applying this componentwise to the exact vector contrast
\eqref{eq:vectorgap} yields Eq.~\eqref{eq:vectorlin}.

Holding $\bar{\bm\Delta}^Q_t$ fixed, the derivative with respect to
$\bm\psi_{t-1}$ is
$J_\lambda=C_0\kappa\lambda B$. Because $B$ is non-negative and row-stochastic,
\[
\|Bx\|_\infty
\le \max_c\sum_dB_{cd}|x_d|
\le\|x\|_\infty,
\]
so $\|B\|_\infty=1$ (equality follows from $B\mathbf1=\mathbf1$). Hence
$C_0\kappa\lambda<1$ makes the linearised map a contraction. Conversely,
$B\mathbf1=\mathbf1$ implies
$J_\lambda\mathbf1=C_0\kappa\lambda\mathbf1$, so the common mode is amplified
when $C_0\kappa\lambda>1$. More generally, if $Bq=\xi q$ then
$J_\lambda q=C_0\kappa\lambda\xi q$, yielding the eigenmode multiplier in
Proposition~\ref{prop:amplification}. Standard spectral properties of
non-negative stochastic matrices are discussed in \citet{HornJohnson2013}.

For the symmetric two-community matrix \eqref{eq:Bgamma}, direct multiplication
gives
\[
Bq_+=q_+,
\qquad
Bq_-=(1-2\Gamma)q_-,
\]
which proves Corollary~\ref{cor:twomodes}. On $0\le\Gamma\le1/2$ the magnitude
of the anti-symmetric eigenvalue falls monotonically from one to zero.

Finally, $0\le\phi_{c,t}(a)<1$ and
$\phi_{c,t}(1)+\phi_{c,t}(2)=N_c^{-1}\sum_{i\in V_c}C_{i,t}<1$, so
$|\psi_{c,t}|\le1$. The linear statement ``amplified'' must therefore be read
as local instability/escape from indifference; nonlinearities and the bounded
state space necessarily govern later motion.

\FloatBarrier
\subsection{Threshold decomposition and comparative statics}
\label{app:thresholdcs}

The common-mode threshold is
$\lambda^\star=\sigma^2/(\beta aC_0)$. Treating $C_0$ as fixed gives the
partial derivatives in Eq.~\eqref{eq:partialcs}. Substituting the explicit
indifference confidence \eqref{eq:C0} yields Eq.~\eqref{eq:lambdastarstructural}.
Let $u=1+a^2/(\sigma^2\tau_0)>1$. Then
\[
\frac{\partial\lambda^\star}{\partial\kappa_1}=0,
\qquad
\frac{\partial\lambda^\star}{\partial\kappa_2}
=\frac{\sigma^2}{\beta a}u^{\kappa_2}\log u>0,
\]
and
\[
\frac{\partial\lambda^\star}{\partial\tau_0}
=-\frac{a\kappa_2}{\beta\tau_0^2}u^{\kappa_2-1}<0.
\]
The total derivatives with respect to $a$ and $\sigma$ contain both a direct
choice-sensitivity term and an indirect $C_0$ term:
\[
\frac{\partial\lambda^\star}{\partial a}
=-\frac{\sigma^2(1+u^{\kappa_2})}{\beta a^2}
+\frac{2\kappa_2}{\beta\tau_0}u^{\kappa_2-1},
\]
\[
\frac{\partial\lambda^\star}{\partial\sigma}
=\frac{2\sigma}{\beta a}
\left[1+u^{\kappa_2}
-\kappa_2\frac{a^2}{\sigma^2\tau_0}u^{\kappa_2-1}\right].
\]
Unlike the partial derivatives holding $C_0$ fixed, these total effects need not
have a parameter-free sign for arbitrary $\kappa_2$.

\section{ODD+D model description and design rationale}
\label{app:odd}
This appendix follows the ODD+D extension for models with human decision-making \citep{MullerBohmeFrankDresslerGroeneveldKlassertMartinSchluterSchulzeWeiseSchwarz2013} and the second ODD update \citep{GrimmRailsbackVincenotEtAl2020}. The latter recommends a concise summary in the journal article alongside a complete description that makes model rationale, evaluation and implementation links explicit. Section~\ref{sec:model} provides that summary; the material below gives the full protocol-level description.

\paragraph{B.1 Purpose and patterns.}
The model's purpose is to determine when credibility-weighted social transmission dampens early differences and when it amplifies them. It is used to study the patterns \textbf{P1} efficient consensus under moderate transmission; \textbf{P2} wrong consensus under strong transmission after an early confident error; \textbf{P3} polarisation under low cross-community permeability; \textbf{P4} a non-monotone effect of social-transmission intensity; \textbf{P5} quotient accuracy away from phase boundaries; and \textbf{P6} quotient degradation near phase boundaries. It is \emph{not} designed to reproduce empirical opinion time series, continuous-action choice, or non-stationary environments.

\paragraph{B.2 Entities, state variables and scales.}
Entities are agents, communities, two arms and the weighted network. Time is
discrete at the trial level with horizon $T$. Agent state: latent values
$Q_{i,t}(a)$, social signal $S_{i,t}(a)$, augmented value $V_{i,t}(a)$, action
$A_{i,t}$, decision time $\tau_{i,t}$, confidence $C_{i,t}$. Community state:
action mass $m_{c,t}(a)$, confidence-weighted mass $\phi_{c,t}(a)$, mean value
$\bar Q_{c,t}(a)$; the network is summarised by the row-stochastic coupling
matrix $B$.

\paragraph{B.3 Process overview and scheduling.}
Each trial executes, synchronously across agents: (1) construct the anticipatory
signal from the previous period; (2) form augmented values; (3) generate the
drift--diffusion choice; (4) draw the decision time; (5) compute confidence;
(6) draw rewards; (7) compute all private and social prediction errors from the
\emph{pre-update} values; (8) apply the private and social updates
\emph{simultaneously}; (9) record observables. The update is synchronous and
simultaneous by construction (tested item~4, Appendix~\ref{app:tests}).

\paragraph{B.4 Design concepts.}
\emph{Basic principles}: reward-prediction-error learning, process-based choice,
process-generated confidence, dual-channel sociality. \emph{Emergence}: regimes and the
phase structure emerge from agent rules and coupling, not from imposed
aggregates. \emph{Adaptation}: value revision via private and social prediction
errors. \emph{Objectives}: agents pursue reward myopically through learned
values; there is no global optimisation. \emph{Learning}: confidence-modulated
private and social learning rates. \emph{Prediction}: agents act on current
augmented values; no forward simulation. \emph{Sensing}: neighbours' previous
actions and confidence (anticipatory) and current outcomes and confidence
(retrospective). \emph{Interaction}: weighted by $W$/$B$ and by credibility.
\emph{Stochasticity}: diffusion noise and Bernoulli rewards. \emph{Collectives}:
communities. \emph{Observation}: the observables of Section~\ref{sec:design}.

\paragraph{B.5 Initialisation.}
Initial values are community-constant; early leads are imposed through the
initial value asymmetry. Community sizes, coupling matrix, reward means, horizon
and seeds are listed per scenario in Table~\ref{tab:scenarios}. Confidence is not
initialised (it is generated at each trial).

\paragraph{B.6 Input data.}
None. The model is a synthetic computational mechanism model and uses no external
empirical data.

\paragraph{B.7 Submodels.}
(1) anticipatory signal \eqref{eq:Si}; (2) augmented value \eqref{eq:V};
(3) DDM choice \eqref{eq:choiceprob} and mean first-passage time \eqref{eq:meanfpt};
(4) reaction-time approximation (Section~\ref{sec:model}, Appendix~\ref{app:tests});
(5) confidence \eqref{eq:confidence}; (6) private learning \eqref{eq:privatelr};
(7) retrospective social learning \eqref{eq:socpe}; (8) projection to $[0,1]$
\eqref{eq:update}; (9) regime classification (Section~\ref{sec:design});
(10) quotient approximation \eqref{eq:quotientfield}. Each maps to a code file
and test in Table~\ref{tab:codemap}.

\paragraph{B.8 Parameters.}
Global default parameters are in Table~\ref{tab:defaults}; per-figure scenario
parameters in Table~\ref{tab:scenarios}.

\paragraph{B.9 Code-location map.}
Table~\ref{tab:codemap} maps each model element to its equation, code file and
test.

\paragraph{B.10 Verification, validation and reproducibility.}
See Appendix~\ref{app:tests} for the test contract and drift--diffusion
validation, and the replication metadata at the end of the main text for exact
commands.

\FloatBarrier
\section{Verification suite and drift--diffusion validation}
\label{app:tests}
The implementation passes the following automated tests (all green in the
replication package): (1) row-stochasticity of $W$ and $B$; (2) confidence
bounded in $(0,1)$; (3) values bounded in $[0,1]$; (4) the simultaneous-update
identity, checked against an independent reference and shown to differ from a
sequential update; (5) seed reproducibility; (6) the no-social baseline, which
is invariant to $B$ when $\lambda=\eta=0$ and learns the better arm; (7)
decoupling of communities when $B=I$; (8) exchangeability under full mixing;
(9) exactness of the quotient field \eqref{eq:quotientfield} under balanced
block weights, checked against explicit weighting; (10) the self-weighting
convention (anticipatory includes self, retrospective excludes self). Several
of these are now analytical consequences rather than merely numerical checks:
Appendix~\ref{app:proofs} proves row-stochasticity under balanced blocks,
confidence and value boundedness, the quotient identity, the $B=I$ and uniform
mixing implications, and the exact $B_{cc}/N_c$ self-exclusion bound. The tests
therefore function as implementation-regression checks against those derived
properties. In addition, the exact choice probability \eqref{eq:choiceprob} and the analytic
mean decision time \eqref{eq:meanfpt} are validated against an
Euler--Maruyama simulation of \eqref{eq:ddm} (the empirical choice frequency
matches \eqref{eq:choiceprob} to within $0.025$ and the empirical mean reaction
time to within $8\%$ across the drift range used), and the fast balanced-block
engine reproduces the dense-$W$ reference engine to machine precision on a
balanced network. Appendix~\ref{app:robust} (check R7) further shows that regime
classification is insensitive to the reaction-time approximation.

\FloatBarrier
\section{Trajectory-level quotient validation}
\label{app:quotient-trajectories}
\begin{figure}[H]
\centering
\includegraphics[width=\textwidth]{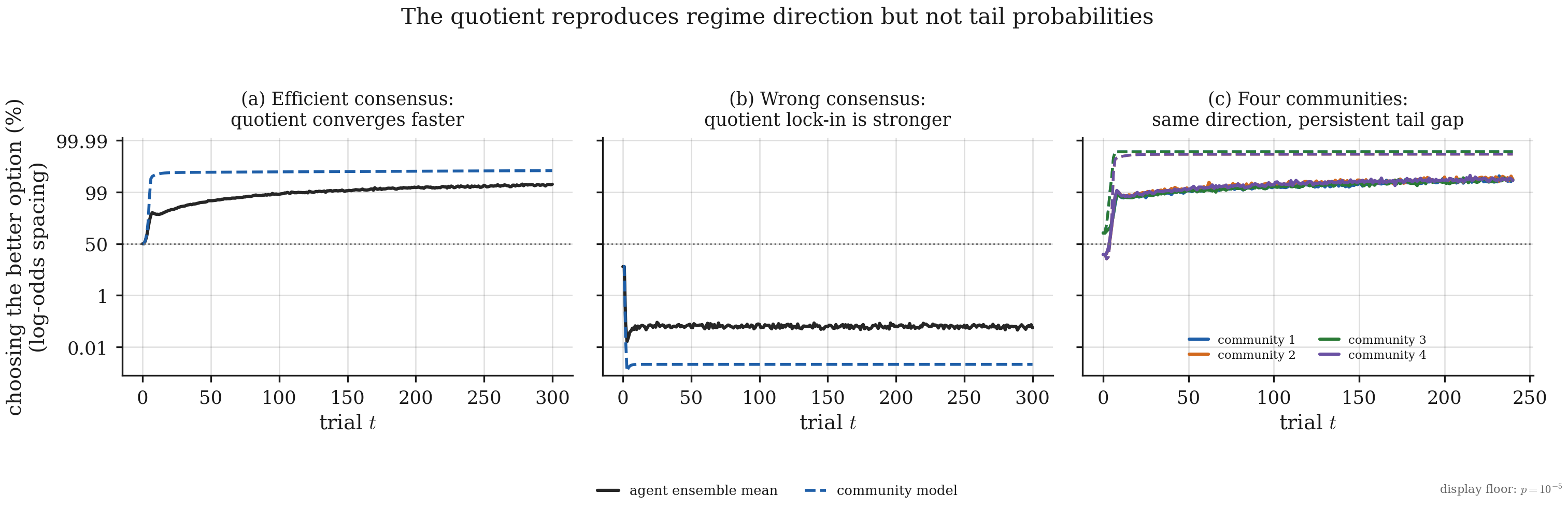}
\caption{Trajectory-level comparison of the agent model and its community
approximation. Solid lines report the agent-ensemble mean and dashed lines the
deterministic quotient; vertical position is the probability of choosing the
better option, displayed with log-odds spacing so differences near zero and one
remain visible. \textbf{(a)} In efficient consensus, both models select the
better arm but the quotient converges faster. \textbf{(b)} In wrong consensus,
both select the worse arm and their aggregate paths remain close.
\textbf{(c)} In the four-community stress test, colours identify communities
and line style identifies model level; both models preserve direction and
community ordering, with a persistent difference in tail probabilities.}
\label{fig:quotient-trajectories}
\end{figure}

Figure~\ref{fig:quotient-trajectories} exposes the trajectories underlying the
error summaries in Figure~\ref{fig:quotient}. It is placed in the appendix
because its purpose is diagnostic: the main-text figure establishes the domain
of validity, while this companion shows how particular discrepancies arise.

\FloatBarrier
\section{Parameters, scenarios and code map}
\label{app:params}
\FloatBarrier
\begin{table}[H]\centering\small
\caption{Global default parameters (\texttt{code/credibility/config.py}). Confidence map is \texttt{decision} (Eq.~\ref{eq:confidence}); values are projected to $[0,1]$ each trial; reaction times use the inverse-Gaussian approximation.}
\label{tab:defaults}
\begin{tabular}{llcl}\toprule
Symbol & Default & Role & Config key \\ \midrule
$\beta$ & 6.0 & value sensitivity (drift scaling) & \texttt{beta} \\
$\sigma$ & 1.0 & internal decision noise & \texttt{sigma} \\
$a$ & 1.0 & speed--accuracy threshold & \texttt{a\_thr} \\
$\kappa_1$ & 3.0 & confidence: evidence weight & \texttt{kappa1} \\
$\kappa_2$ & 1.0 & confidence: decision-time weight & \texttt{kappa2} \\
$\tau_0$ & 0.5 & reaction-time scale & \texttt{tau0} \\
$\alpha^{\min}$ & 0.05 & min reward learning rate & \texttt{alpha\_min} \\
$\alpha^{\max}$ & 0.4 & max reward learning rate & \texttt{alpha\_max} \\
$\gamma$ & 0.5 & retrospective credit-assignment scale & \texttt{gamma} \\
$\omega$ & 1.0 & social learning-rate exponent & \texttt{omega} \\
$\varepsilon$ & 0.001 & $\bar C$ denominator floor & \texttt{eps\_soc} \\
RT disp. & 0.3 & reaction-time squared CV & \texttt{rt\_dispersion} \\
\bottomrule\end{tabular}
\end{table}

\FloatBarrier
\begin{table}[H]\centering\footnotesize
\caption{Scenario parameters for all reported figures (standard scale). $N$ is total agents, $T$ horizon, $R$ replications, $\Delta\mu=\mu_1-\mu_2$, $\lambda$ anticipatory weight, $\eta$ retrospective weight, $\Gamma=1-B_{cc}$ permeability. Regime threshold $0.9$ throughout. Configuration: \texttt{code/credibility/scenarios.py}.}
\label{tab:scenarios}
\resizebox{\textwidth}{!}{\begin{tabular}{lcccccccl}\toprule
Scenario & $N$ & $T$ & $R$ & $\Delta\mu$ & $\lambda$ & $\eta$ & $\Gamma$ & init. lead \\ \midrule
Fig.~2 efficient & 400 & 300 & 300 & 0.20 & 0.40 & 0.30 & 0.30 & neutral \\
Fig.~2 wrong & 400 & 300 & 300 & 0.10 & 0.90 & 0.20 & 0.30 & asym. \\
Fig.~2 polarised & 400 & 300 & 300 & 0.10 & 0.90 & 0.25 & 0.02 & asym. \\
Fig.~3 phase ($\lambda,\Gamma$ swept) & 160 & 240 & 120 & 0.10 & [0,1.6] & 0.25 & [.01,.5] & asym. \\
Fig.~4 ablation point & 400 & 260 & 300 & 0.10 & 0.60 & 0.30 & 0.15 & asym. \\
Fig.~5 confidence & 300 & 260 & 60 & 0.10 & 0.50 & 0.25 & 0.15 & neutral \\
Fig.~6 four-community & 320 & 240 & 250 & 0.10 & 0.60 & 0.25 & 0.30 & asym. \\
\bottomrule\end{tabular}}
\end{table}

\FloatBarrier
\begin{table}[H]\centering\small
\caption{Equation-to-code-to-test map. Code paths are under \texttt{code/credibility/}; tests under \texttt{code/tests/test\_credibility.py}.}
\label{tab:codemap}
\begin{tabular}{llll}\toprule
Model element & Equation & Code & Test \\ \midrule
Anticipatory signal & Eq.~\ref{eq:Si} & \texttt{model.py: anticipatory\_field\_block} & \texttt{test\_quotient\_field\_exact} \\
Augmented value & Eq.~\ref{eq:V} & \texttt{model.py: simulate\_blocks} & \texttt{test\_block\_dense\_*} \\
DDM choice & Eq.~\ref{eq:choiceprob} & \texttt{ddm.py: choice\_prob\_up} & \texttt{test\_ddm\_choice\_*} \\
Reaction time & \S\ref{sec:model} & \texttt{ddm.py: sample\_decision\_time} & \texttt{test\_ddm\_mean\_fpt\_*} \\
Confidence & Eq.~\ref{eq:confidence} & \texttt{confidence.py} & \texttt{test\_confidence\_bounded} \\
Private learning & Eq.~\ref{eq:privatelr} & \texttt{model.py: \_private\_lr} & \texttt{test\_simultaneous\_*} \\
Retrospective learning & Eq.~\ref{eq:socpe} & \texttt{model.py: value\_update\_*} & \texttt{test\_self\_weighting\_*} \\
Bounded values & Eq.~\ref{eq:update} & \texttt{model.py: np.clip} & \texttt{test\_values\_bounded} \\
Regime classification & \S\ref{sec:design} & \texttt{metrics.py: classify\_regime} & \texttt{test\_regime\_classification} \\
Quotient (meso) & Eq.~\ref{eq:quotientfield} & \texttt{model.py: simulate\_meso} & \texttt{test\_meso\_*} \\
Uncertainty & \S\ref{sec:identification} & \texttt{uncertainty.py} & \texttt{---} \\
\bottomrule\end{tabular}
\end{table}

\FloatBarrier

\section{Confidence-modulated learning and transmission}
\label{app:plasticity}
\begin{figure}[H]
\centering
\includegraphics[width=0.92\textwidth]{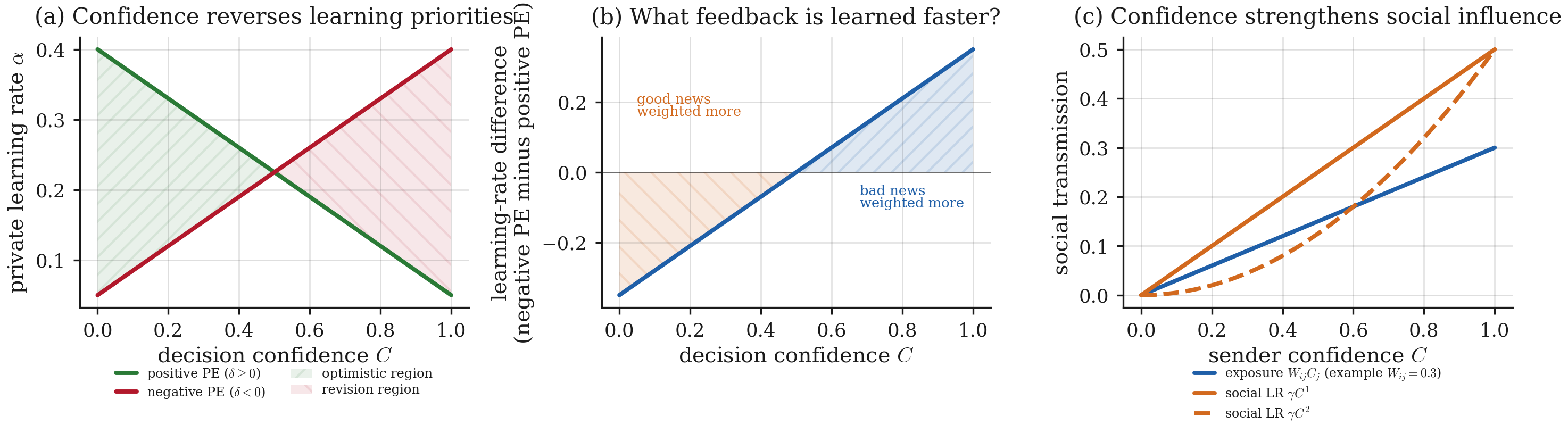}
\caption{Quantitative companion to Figure~\ref{fig:plasticity}. \textbf{(a)}~the
exact confidence-gated learning rates for good news ($\alpha_{\max}-(\alpha_{\max}-\alpha_{\min})C$)
and bad news ($\alpha_{\min}+(\alpha_{\max}-\alpha_{\min})C$), crossing at
$C=\tfrac12$. \textbf{(b)}~An illustrative normalised relative-confidence index $C/(C+\bar C)$ and the social
learning rate $\gamma C^{\omega}$ ($\omega=1,2$) versus sender confidence. The operative anticipatory exposure in the model is \(W_{ij}C_{j,t}\) as defined in Eq.~\eqref{eq:Si}.}
\label{fig:plasticityquant}
\end{figure}

Figure~\ref{fig:plasticity} in the main text is a schematic; the exact
relationships behind it are plotted here. The private learning rate
\eqref{eq:privatelr} is valence-asymmetric in confidence
(Figure~\ref{fig:plasticityquant}a): the good-news and bad-news rates cross at
$C=\tfrac12$, so a low-confidence agent weights success and a high-confidence
agent weights failure. The same confidence also enters social transmission. Figure~\ref{fig:plasticityquant}b plots an illustrative normalised relative-confidence index $C/(C+\bar C)$ alongside the social learning rate $\gamma\,C^{\omega}$; the operative anticipatory exposure remains $W_{ij}C_{j,t}$. These are the intra-agent and inter-agent roles of the same process-generated confidence signal.

\FloatBarrier
\section{Robustness}
\label{app:robust}
\begin{figure}[H]
\centering
\includegraphics[width=\textwidth]{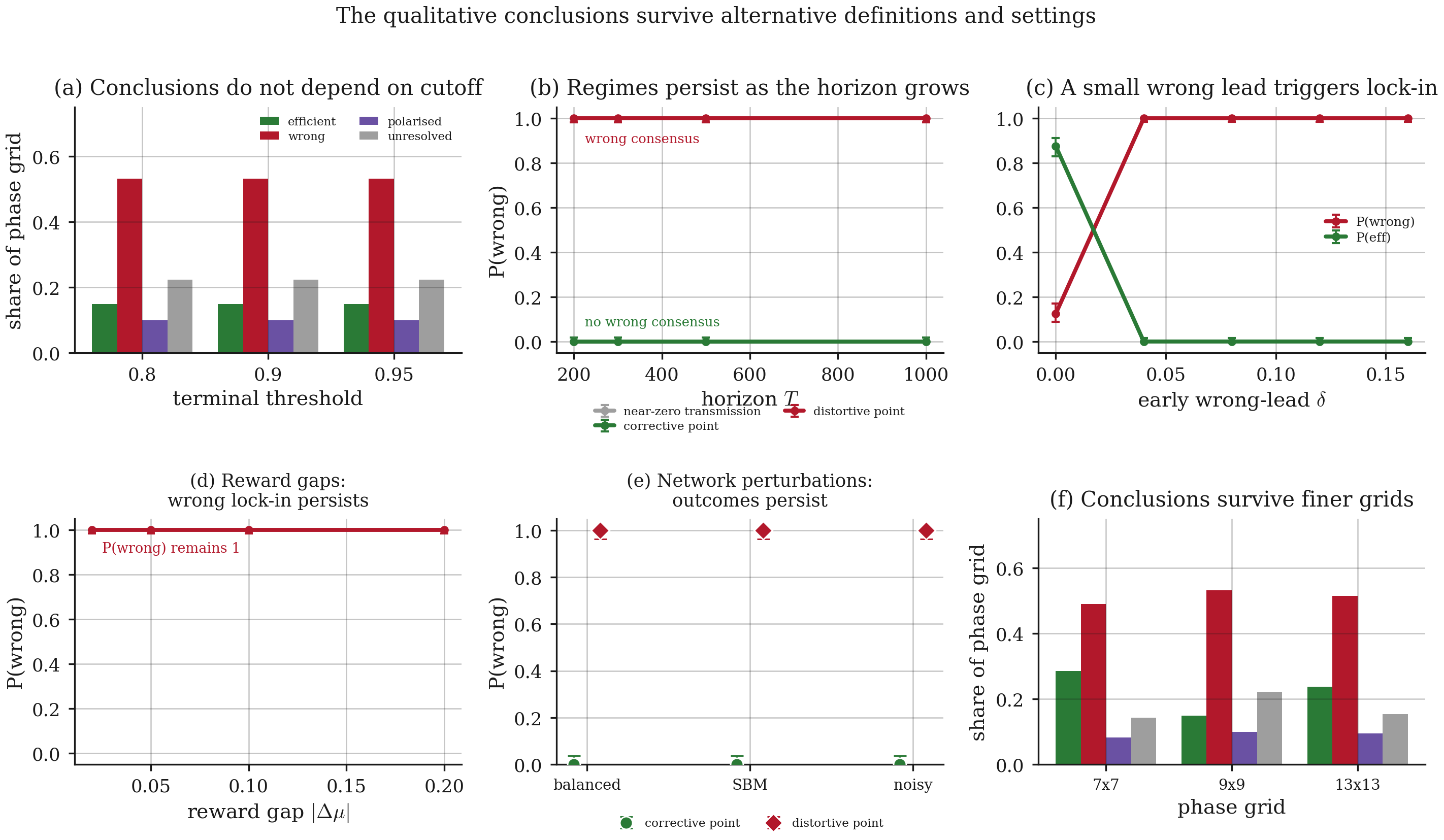}
\caption{Robustness of the phase structure. (a) regime area fractions are
invariant to the terminal threshold; (b) P(wrong) versus horizon for three
operating points; (c) P(wrong) and P(efficient) versus early-lead strength
$\delta$; (d) P(wrong) versus reward gap with Wilson intervals; (e) P(wrong)
under balanced, weighted-SBM and noisy graphs; (f) regime area fractions versus
phase-grid resolution. Full tables in Table~\ref{tab:robustness}.}
\label{fig:robustness}
\end{figure}

We probe seven sensitivities (full CSVs in \texttt{paper/tables/};
Table~\ref{tab:robustness}, Figure~\ref{fig:robustness}). \textbf{R1} terminal
threshold $\{0.80,0.90,0.95\}$: dominant-regime area fractions are invariant.
\textbf{R2} horizon $T\in\{200,300,500,1000\}$: the unresolved region at
$\lambda\to0$ persists (genuine, not slow convergence); corrective and distortive
points are stable. \textbf{R3} early-lead strength $\delta$: a small lead
($\delta=0.04$) already triggers wrong consensus at the distortive point.
\textbf{R4} reward gap $|\Delta\mu|\in\{0.02,\dots,0.20\}$: wrong consensus
persists at the distortive point even for clearly distinguishable arms.
\textbf{R5} random graphs: balanced block, weighted stochastic-block and noisy
row-normalised $W$ give the same regimes, so the results do not depend on exact
quotient symmetry. \textbf{R6} phase-grid resolution $\{7^2,9^2,13^2\}$: the
area fractions are stable. \textbf{R7} reaction-time approximation: the
inverse-Gaussian and deterministic-mean reaction times give nearly identical
regime maps.

\FloatBarrier
\begin{table}[H]\centering\small
\caption{Robustness summary (Appendix~\ref{app:robust}). Entries are the dominant-regime area fractions of the $(\lambda,\Gamma)$ plane, or P(wrong) at the operating point, as indicated. Full tables: \texttt{paper/tables/robustness\_*.csv}.}
\label{tab:robustness}
\begin{tabular}{lll}\toprule
Check & Variation & Result \\ \midrule
R1 threshold & $0.80/0.90/0.95$ & wrong-area $=0.53/0.53/0.53$ (invariant) \\
R2 horizon & $T=200\!-\!1000$ & near-zero stays unresolved; corrective/distortive stable \\
R3 initial lead & $\delta=0\!-\!0.16$ & P(wrong) $=0.12\!\to\!1.00$ (small lead suffices) \\
R4 reward gap & $|\Delta\mu|=0.02\!-\!0.20$ & P(wrong) $=1.00\!-\!1.00$ (persists) \\
R5 random graphs & balanced/SBM/noisy & distortive P(wrong) $=1.00/1.00/1.00$ \\
R6 grid resolution & $7^2/9^2/13^2$ & wrong-area $=0.49/0.53/0.51$ \\
R7 RT approximation & inverse-Gaussian vs.\ mean & wrong-area $=0.49$ vs.\ $0.49$ \\
\bottomrule\end{tabular}
\end{table}

\FloatBarrier

\FloatBarrier
\section{Stochastically generated early-lead test of the amplification threshold}
\label{app:earlylead}
\begin{figure}[H]
\centering
\includegraphics[width=\textwidth]{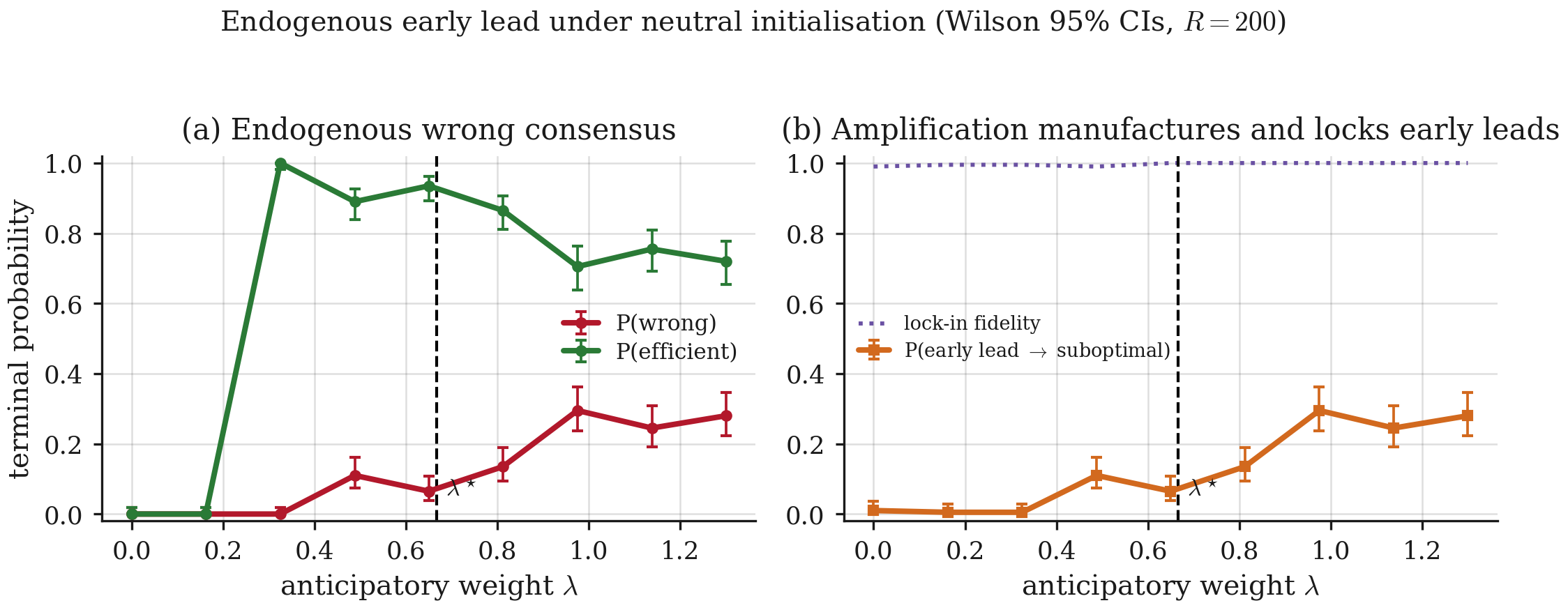}
\caption{Stochastically generated early lead (neutral initialisation, $R=200$ per $\lambda$).
(a) P(wrong) is $\approx0$ below the local amplification threshold
$\lambda^\star\approx0.67$ and rises above it, mirrored by a fall in
P(efficient). (b) the frequency of early suboptimal leads rises with $\lambda$
and they are locked in with fidelity $\approx1$. Data:
\texttt{tables/endogenous\_early\_lead.csv}.}
\label{fig:earlylead}
\end{figure}

Proposition~\ref{prop:amplification} is a \emph{local} statement: it predicts when
small confidence-weighted mass gaps become self-amplifying, not the global regime
map. We test its predictive content without any imposed lead. Agents start from
neutral values ($Q=0.5$ on both arms) with a small reward gap
($\mu=0.53/0.47$) in two symmetric communities; a $25$-trial burn-in lets a
stochastic early asymmetry emerge, after which we measure the population
confidence-weighted mass gap and the terminal regime ($R=200$ per $\lambda$).
Figure~\ref{fig:earlylead} shows that wrong consensus also arises from neutral initial values. It is essentially absent below the neighbourhood of the local threshold and increases as $\lambda$ enters the amplifying region around $\lambda^\star\approx0.67$. Lock-in fidelity is $\approx1$: conditional on a terminal consensus, the selected arm almost always matches the stochastic burn-in lead. Thus the analytical threshold has predictive content for the onset of amplification even when the initial asymmetry is generated by early stochastic outcomes, and the wrong-consensus result does not depend on a hand-picked initial lead.

\clearpage
\bibliographystyle{apalike}
\bibliography{Bontemps_Banerjee_2026_decision_generated_credibility}

\end{document}